%% file: main.tex
\documentclass[letterpaper, 10 pt, conference]{ieeeconf}  
\IEEEoverridecommandlockouts                              

\usepackage[utf8]{inputenc} 
\usepackage{graphics}   
\usepackage{amssymb, amsmath}   

\usepackage{amsthm}     
\usepackage{microtype}  
\usepackage{epstopdf}   
\usepackage{url}    
\usepackage[noadjust]{cite} 
\usepackage{overpic}    
\usepackage[linesnumbered,ruled,vlined]{algorithm2e}    
\usepackage[dvipsnames]{xcolor} 
\usepackage{subfig}
\usepackage{wrapfig}
\usepackage[font=footnotesize,labelfont=bf]{caption}
\usepackage{multirow}   
\usepackage{footnote}   
\usepackage{soul}   
\usepackage{tikz}   
\usetikzlibrary{tikzmark}

\usepackage[textsize=scriptsize]{todonotes}

\makeatletter
\let\NAT@parse\undefined
\makeatother
\usepackage[hidelinks]{hyperref}   
\newtheorem{problem}{Problem}

\theoremstyle{definition}

\theoremstyle{remark}

\DeclareMathOperator*{\argmax}{arg\,max}

\newcommand{\pag}{\textsc{PaG}\xspace}
\newcommand{\dipn}{\textsc{DIPN}\xspace}
\newcommand{\gn}{\textsc{GN}\xspace}

\newcommand{\dipngn}{\textsc{DIPN+GN}\xspace}

\font\titlefont=ptmb at 14.7pt
\title{\titlefont
DIPN: Deep Interaction Prediction Network with Application to Clutter Removal
}
\author{Baichuan Huang$^{1}$, Shuai D. Han$^{1}$, Abdeslam Boularias$^{1}$, and 
Jingjin Yu$^{1}$
\thanks{$^{1}$B. Huang, S. D. Han, A. Boularias, and J. Yu are with the Department of 
Computer Science, Rutgers, the State University of New Jersey, Piscataway, NJ, USA. 
Emails: {\tt\small \{baichuan.huang, shuai.han, abdeslam.boularias, jingjin.yu\}@rutgers.edu}.
This work is supported in part by NSF awards IIS-1845888, IIS-1734492, IIS-1846043, and CCF-1934924.
}%
}

\begin{document}

\maketitle
\thispagestyle{empty}
\pagestyle{empty}

\begin{abstract}
We propose a Deep Interaction Prediction Network (\dipn) for learning to predict  
complex interactions that ensue as a robot end-effector pushes multiple objects, 
whose physical properties, including size, shape, mass, and friction coefficients 
may be unknown \emph{a priori}. \dipn ``imagines'' the effect of a push action and 
generates an accurate synthetic image of the predicted outcome. \dipn is shown to 
be sample efficient when trained in simulation or with a real robotic system. The 
high accuracy of \dipn allows direct integration with a grasp network, yielding a 
robotic manipulation system capable of executing challenging clutter removal tasks 
while being trained in a fully self-supervised manner. The overall network
demonstrates intelligent behavior in selecting proper actions between push and 
grasp for completing clutter removal tasks and significantly outperforms the 
previous state-of-the-art. Remarkably, \dipn achieves even better performance on 
the real robotic hardware system than in simulation.
Videos, code, and experiments log are available at \href{https://github.com/rutgers-arc-lab/dipn}{\texttt{\textcolor{OrangeRed}{https://github.com/rutgers-arc-lab/dipn}}}. 
\end{abstract}

\section{Introduction}\label{sec:introduction}

\begin{figure}[ht!]
    \centering
    \subfloat[]{
        \includegraphics[height=1.5in]{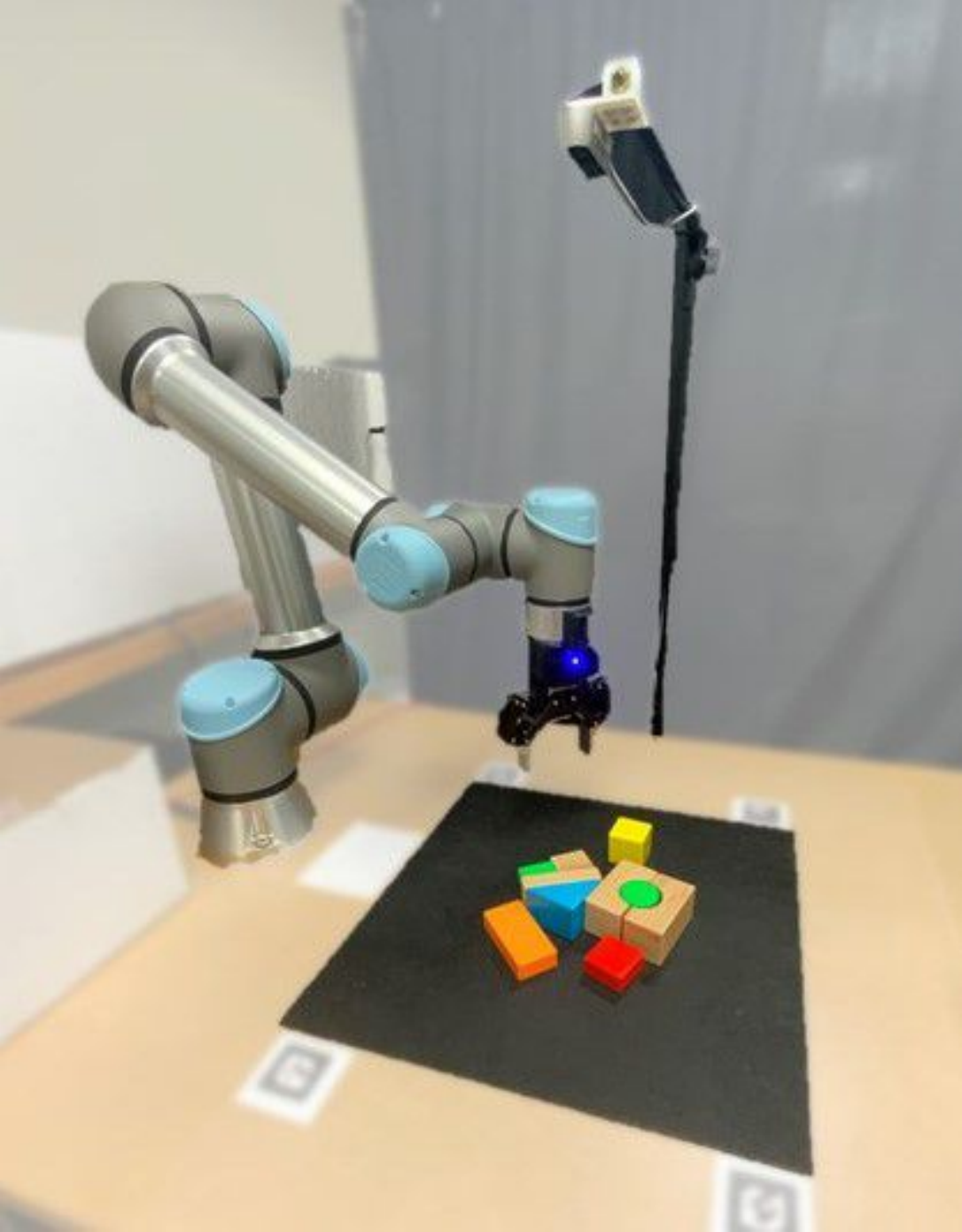}\label{fig:system-setup}
    }
    \hfill
    \subfloat[]{
        \includegraphics[height=1.5in]{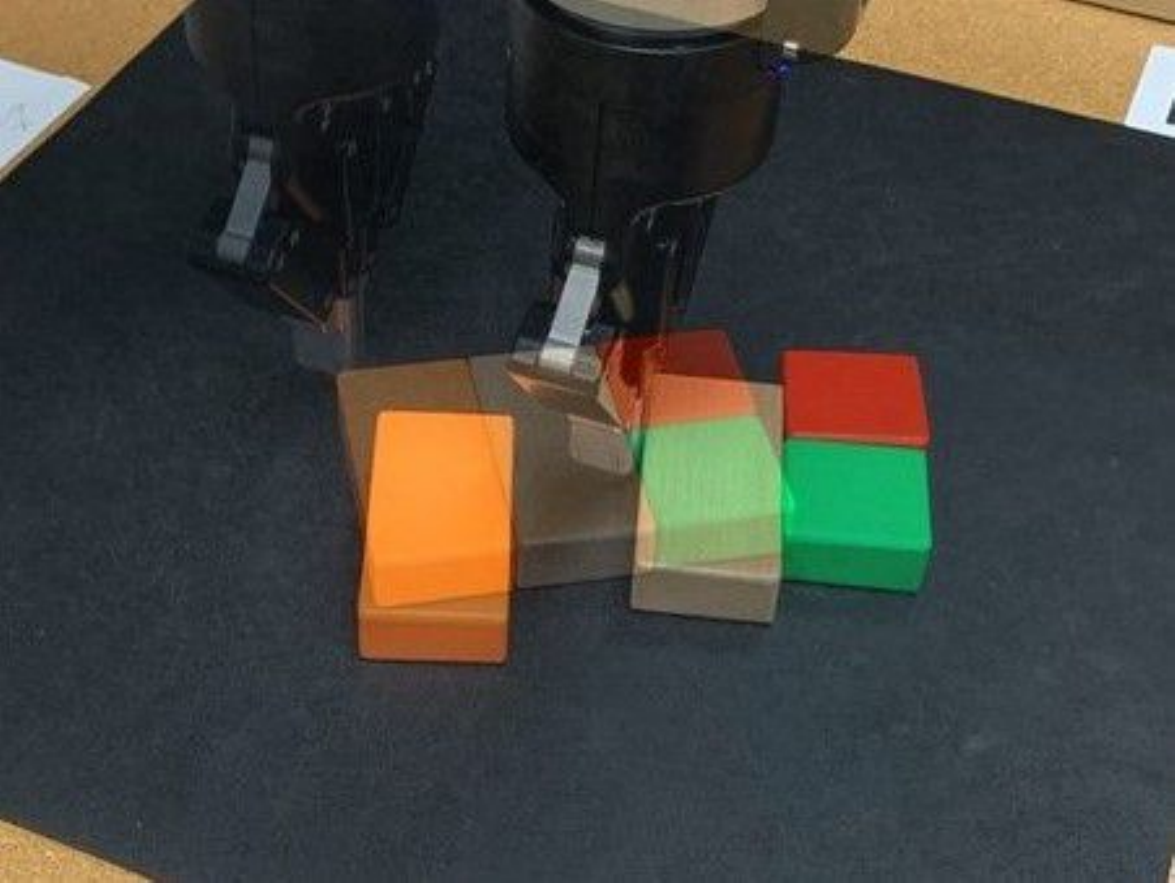}\label{fig:push-action-example}
    }
    
    
    \subfloat[]{
        \begin{overpic}[width=\linewidth]{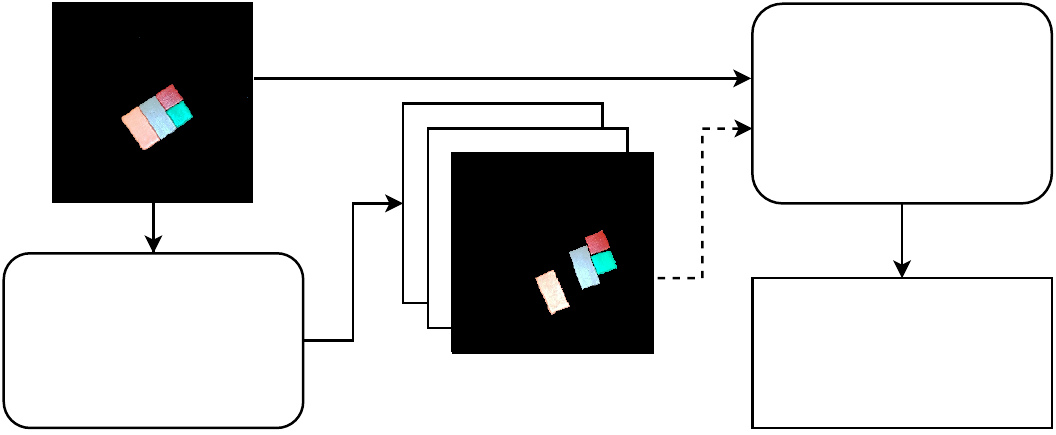}
            \footnotesize
            \put(24.5, 38) {State observation}
            \put(14.5, 13) {\makebox(0,0){Deep Interaction}}
            \put(14.5, 9) {\makebox(0,0){Prediction Network}}
            \put(14.5, 4.5) {\makebox(0,0){(\dipn)}}
            \put(50, 4) {\makebox(0,0){Predicted states after push}}
            \put(85.25, 35) {\makebox(0,0){Deep}}
            \put(85.25, 31) {\makebox(0,0){Grasp Network}}
            \put(85.25, 26.5) {\makebox(0,0){(\gn)}}
            \put(85.25, 9.5) {\makebox(0,0){Grasp or push}}
            \put(85.25, 6) {\makebox(0,0){action}}
        \end{overpic}
        \label{fig:system-architecture} 
    }
    \caption{\label{fig:intro} 
    (a) The system setup includes a workspace with objects to remove, a Universal Robots UR-5e manipulator with a Robotiq 2F-85 two-finger gripper, and an Intel RealSense D435 RGB-D camera. 
    (b) An example push action and superimposed images of scenes before and after the push. 
    (c) System architecture of our pipeline, and one predicted image that \dipn can generate for the push shown in (b). Notice the similarity between the predicted synthetic image and the real image resulting from the push action.
    }
\end{figure}
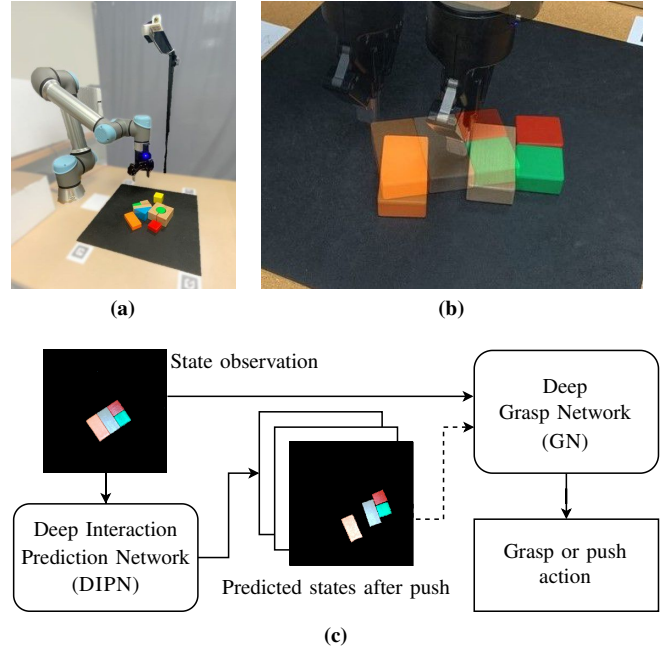

We propose a Deep Interaction Prediction Network (\dipn) for learning object 
interactions directly from examples and using the trained network for accurately 
predicting the poses of the objects after an arbitrary push action (Fig.~\ref{fig:intro}). 
To demonstrate its effectiveness, 
We integrate \dipn with a deep Grasp Network (\gn) for completing challenging 
clutter removal manipulation tasks. 
Given grasp and push actions to choose from, the objective is to remove all objects 
from the scene/workspace with a minimum number of actions. 
In an iteration of push/pick selection (Fig.~\ref{fig:system-architecture}), the 
system examines the scene and samples a large number of candidate grasp and push 
actions. Grasps are immediately scored by \gn, whereas for each candidate push 
action, \dipn generates an image corresponding to the predicted outcome. In a sense, 
\dipn ``imagines'' what happens to the current scene if the robot executes a 
certain push. The predicted future images are also scored by \gn; the action with 
the highest expected score, either a push or a grasp, is then executed. 

Our extensive evaluation demonstrates that \dipn can accurately predict objects' 
poses after a push action with collisions, resulting in less than $10\%$ average 
single object pose error in terms of IoU (Intersection-over-Union), a significant 
improvement over the compared baselines.
Push prediction by \dipn generates clear synthetic images that can be used by \gn 
to evaluate grasp actions in future states. Together with \gn, our entire pipeline 
achieves $34\%$ higher  completion rate, $20.9\%$ higher grasp success rate, and 
$30.4\%$ higher action efficiency in comparison to~\cite{zeng2018learning} on 
challenging clutter removal scenarios. 
Moreover, experiments suggest that \dipn can learn from randomly generated scenarios 
with the learned policy maintaining high levels of performance on challenging tasks 
involving previously unseen objects.
Remarkably, \dipngn achieves an even better performance on real robotic hardware 
than it does in simulation, within which it was developed.


\section{Related Work}\label{sec:related}
Prehensile (e.g., grasping) and non-prehensile (e.g., pushing) manipulation techniques 
are typically studied separately in robotics
\cite{
    10.1109/TRO.2013.2289018, 
    grasping,
    liang2019pointnetgpd,
    doi:10.1177/0278364912442972,
    Detry2013,Lenz2013,7139793,Yan-2018-113286,DBLP:conf/iccv/MousavianEF19, 
    Pas2015UsingGT,DBLP:conf/icra/PintoG16,pmlr-v78-mahler17a,mahler2017dexnet,kalashnikov2018qtopt, 
    10.3389/frobt.2020.00008, 
    Lynch-1993-15932,Mason86,Mason96,Mason99,Mason98,Mason-1985-15649, 
    doi:10.1177/027836499601500603, 
    Zhou2016,JJZhou2018,DBLP:journals/ijrr/ZhouHM19, 
    han2018complexity,
    DBLP:conf/rss/DogarS11, 
    DBLP:conf/icra/BauzaR17, 
    NIPS2016_6161,7989023,watters2017visual, 
    szegedy2020rearrangement,shome2021fast,
    sergey2015learning,10.5555/2946645.2946684,DBLP:conf/icra/FinnL17,DBLP:journals/corr/GhadirzadehMKB17,ChangkyuIROS2019,PackingICRA2019
}. Seeking to reap the benefit from the apparent synergy between 
push and grasp, recent years have seen increasing interests in using both to tackle challenging 
manipulation problems, including pre-grasp
push~\cite{Kaiyu2019,Dogar2010PushgraspingWD,6631288,Dogar_2012_7076,King2013PregraspMA} 
and more recently push-assisted grasping~\cite{zeng2018learning}. 

\input{related}


\section{Problem Formulation}\label{sec:preliminaries}

We denote the clutter removal problem (Fig.~\ref{fig:system-setup}) as Pushing 
Assisted Grasping (\pag). In a \pag, the workspace of the manipulator is a 
square region containing multiple objects and the volume directly above it. 
A camera is placed on top of the workspace for state observation. Given camera 
images, all objects must be removed using two basic motion primitives, grasp, 
and push, with a minimum number of actions.

In our experimental setup, the workspace has a uniform background color and the 
objects have different shapes, sizes, and colors. The end-effector is a 2-finger 
gripper with a narrow stroke that is slightly larger than the smallest dimension 
of individual objects. Objects are removed one by one, which requires a sequence 
of push and grasp actions.
When deciding on the next action, a state observation is given as an 
RGB-D image, re-projected orthographically, cropped to the workspace's boundary, 
and down-sampled to $224 \times 224$. 

From the down-sampled image, a large set of candidate actions is generated by 
considering each pixel in the image as a potential center of a grasp action or 
initial contact point of a push action.
A grasp action $a^{\text{grasp}} = (x, y, \theta)$ is a vertical top-down grasp 
centered at pixel position $(x, y)$ with the end-effector rotation set to $\theta$
around the vertical axis of the workspace; a grasped object  is subsequently 
transferred outside of the workspace and removed from the scene. Similarly, a 
push action $a^{\text{push}} = (x, y, \theta)$ is a horizontal sweep motion that 
starts at $(x, y)$ and proceeds along $\theta$ direction for a constant distance.
The orientation $\theta$ can be one of $16$ values evenly distributed between $0$ 
and $360$ degrees. That is, the entire action space includes $2 \times 224 \times 
224 \times 16$ different grasp/push actions. 

The problem studied in this paper is defined as: 
\begin{problem}
{\normalfont \textbf{Pushing Assisted Grasping (\pag)}.} 
Given objects in clutter and under the described system setup, choose a sequence 
of push and grasp actions for removing all the objects, based only on images of 
the workspace, while minimizing the number of actions taken.
\end{problem}


\section{Methodology}\label{sec:method} 

We describe the Deep Interaction Prediction Network (\dipn), the Grasp Network (\gn), 
and the integrated pipeline for solving \pag challenges.

\subsection{Deep Interaction Prediction Network (\dipn)}\label{subsec:pp}
\begin{figure*}[ht!]
    \centering
    \begin{overpic}[width=\linewidth]{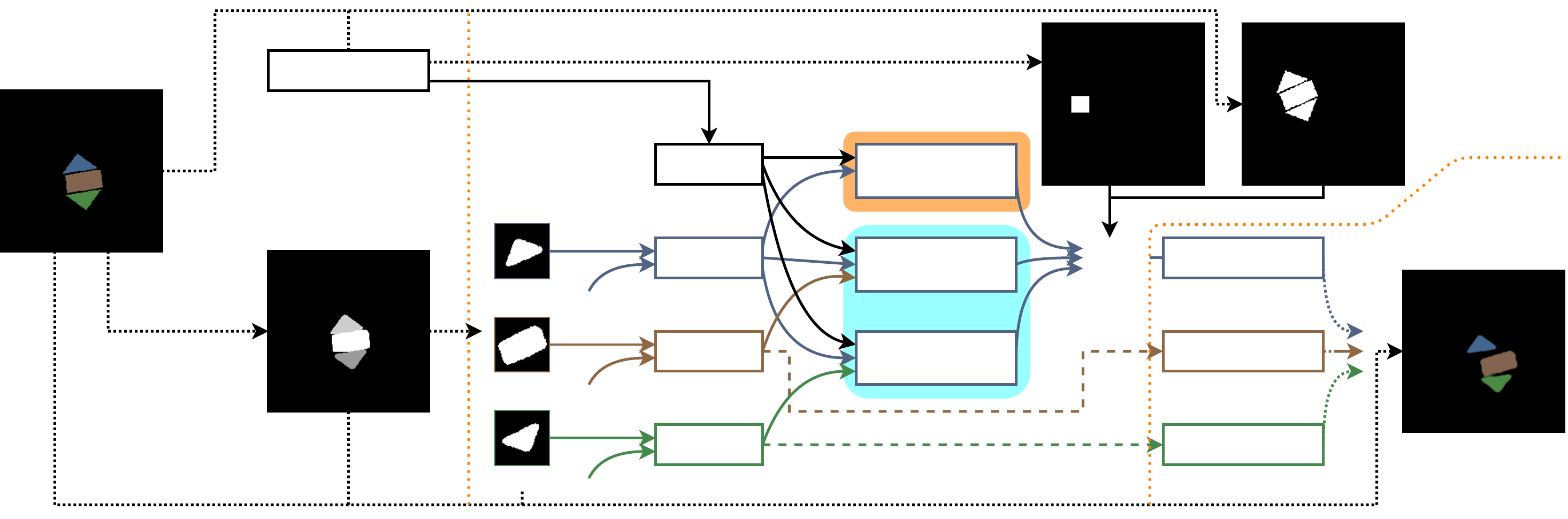}
        \scriptsize
        \put(0, 29.5) {Input image $I$}
        \put(0, 28) {$224 \times 224 \times 3$}
        \put(17, 19) {Segmentation}
        \put(17, 17.5) {$224 \times 224 \times 1$}
        \put(89.5, 18) {Output image $\hat I$}
        \put(89.5, 16.5) {$224 \times 224 \times 4$}
        \put(47.5, 27.5) {Binary push action image $M_p$}
        \put(54.5, 26) {$224 {\times} 224 {\times} \{0{,} 1\}$}
        \put(90, 27.5) {Binary image $M_I$}
        \put(73.5, 29.5) {\textcolor{orange}{\textbf{$M_p$}}}
        \put(86.5, 29.5) {\textcolor{orange}{\textbf{$M_I$}}}
        \put(90, 26) {$224 {\times} 224{\times} {\{}0{,} 1{\}}$}
        \put(22.3, 28.5) {\makebox(0,0){Push action $p$}}
        \put(22.3, 26.25) {\makebox(0,0){\scriptsize $(83, 140, 22.5^\circ)$}}
        \put(7, 10) {Mask R-CNN}
        \put(33.5, 21.5) {\makebox(0,0){\parbox{1in}{\centering Mask image$\{M_i\}$, \\ $60 {\times} 60 {\times} \{0{,} 1\}$, \\ center position $\{c_i\}$}}}
        \put(33.5, 14.4) {\makebox(0,0){\scriptsize $(54, 69)$}}
        \put(33.5, 8.4) {\makebox(0,0){\scriptsize $(67, 90)$}}
        \put(33.5, 2.4) {\makebox(0,0){\scriptsize $(73, 112)$}}
        \put(35.2, 15.8) {\textcolor{orange}{\textbf{$M_1$}}}
        \put(35.2, 9.6){\textcolor{orange}{\textbf{$M_2$}}}
        \put(35.2, 3.8) {\textcolor{orange}{\textbf{$M_3$}}}
        \put(36, 17.5) {\scriptsize ResNet\textsubscript{2}}
        \put(41, 26) {\scriptsize MLP\textsubscript{1}}
        \put(38.5, 14.5) {\scriptsize MLP\textsubscript{2}}
        \put(52.5, 24.4) {\scriptsize MLP\textsubscript{3}}
        \put(52.5, 18.4) {\scriptsize MLP\textsubscript{4}}
        \put(52.5, 12.4) {\scriptsize MLP\textsubscript{4}}
        \put(42.7, 22) {Encoded}
        \put(42.7, 16) {Encoded}
        \put(42.7, 10.1) {Encoded}
        \put(42.7, 4.1) {Encoded}
        \put(59.8, 22.2) {\makebox(0,0){\parbox{1in}{\scriptsize \centering Direct \\ transformation}}}
        \put(59.8, 16.2) {\makebox(0,0){\parbox{1in}{\scriptsize \centering Interactive \\ transformation}}}
        \put(59.8, 10.2) {\makebox(0,0){\parbox{1in}{\scriptsize \centering Interactive \\ transformation}}}
        \put(66, 13.5) {\scriptsize Add}
        \put(73.7, 19.5) {\makebox(0,0){\scriptsize ResNet\textsubscript{1}}}
        \put(69.5, 16.2) {\scriptsize MLP\textsubscript{5}}
        \put(79.3, 16.7) {\makebox(0,0){\scriptsize Transformation}}
        \put(79.3, 14.5) {\makebox(0,0){\scriptsize $(0, 0, 0^{\circ})$}}
        \put(79.3, 10.7) {\makebox(0,0){\scriptsize Transformation}}
        \put(79.3, 8.5) {\makebox(0,0){\scriptsize $(18, -3, 6^{\circ})$}}
        \put(79.3, 4.7) {\makebox(0,0){\scriptsize Transformation}}
        \put(79.3, 2.5) {\makebox(0,0){\scriptsize $(15, 1, -3^{\circ})$}}
    \end{overpic}
    \caption{\label{fig:push-prediction-flowchart} 
    \dipn flow with an example. 
    The network components dedicated to an object are colored with the same as the object. 
    We only show the full network for the blue triangle object; the instance-specific 
    structures for the other objects share the same weights and are simplified as dashed lines. Components inside the orange dot line are the core of the DIPN. The output image is synthesized by applying the predicted transformations to the segments.
    }
\end{figure*}
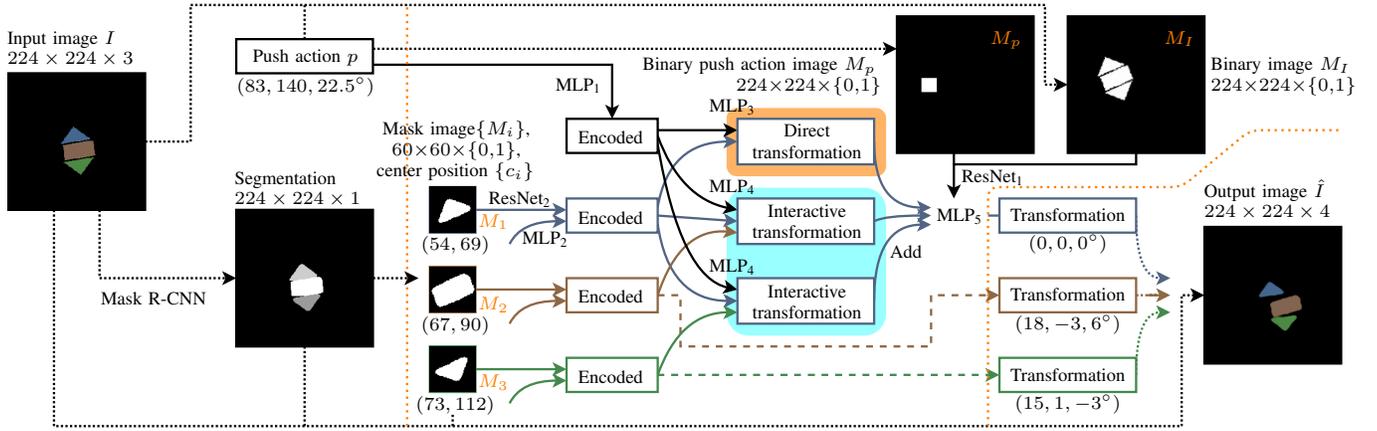

The architecture of our proposed \dipn is outlined  in 
Fig.~\ref{fig:push-prediction-flowchart}.
At a high level, given an image and a candidate push action as inputs, \dipn segments 
the image and then predicts 2D transformations (translations and rotations) for all objects,
and particularly for those affected by the push action, directly or indirectly through a 
cascade of object-object interactions. 
A predicted image of the post-push scene is synthesized by applying the predicted 
transformations on the segments.
We opted against an end-to-end, pixel-to-pixel method as such methods (e.g.,
\cite{NIPS2016_6161}) often lead to blurry or fragmented images, which are not conducive 
to predicting the quality of a potential future grasp action. 

\textbf{Segmentation.} 
\dipn employs Mask R-CNN~\cite{he2017mask} for object segmentation (instance level only, 
without semantic segmentation). The resulting binary masks ($m_i$) and their centers 
($c_i$), one per object, serve as the input to the push prediction module of \dipn. 
Our Mask R-CNN setup has two classes, one for the background and 
one for the objects. 
The network is trained from scratch in a \emph{self-supervised} manner without any human 
intervention: objects are randomly dropped into the workspace, and data is automatically 
collected.
Images that can be easily segmented into separate instances based on color/depth 
information (distinct color blobs) are automatically labeled by the system as 
single instances for training the Mask R-CNN. 
The self-trained Mask R-CNN can then accurately find edges in images of tightly packed 
scenes and even in scenes with novel objects. Note that the data used for training the 
segmentation module are also counted in our evaluation of the data efficiency of our 
technique and the comparisons to alternative techniques.

\begin{wrapfigure}[10]{r}{1.5in}
  \vspace*{-3mm}
  \includegraphics[width=1.5in]{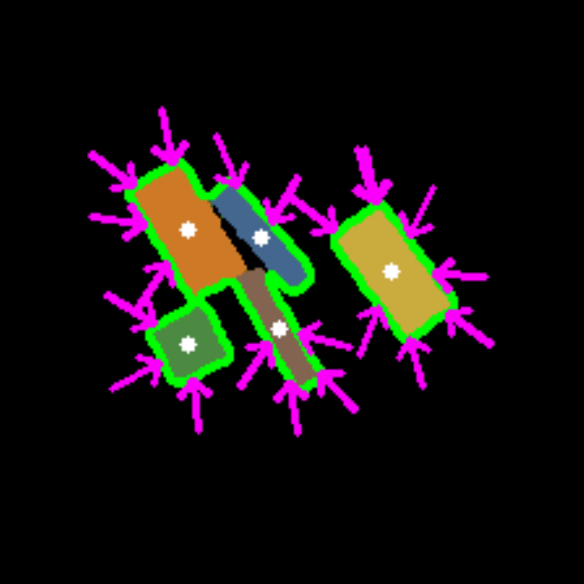}
\end{wrapfigure}
\textbf{Push sampling.}
Based on foreground segmentation,
candidate push actions are generated by uniformly sampling push contact locations 
on the contour of object bundles (see the figure to the right). Push directions point 
to the centers of the objects. Pushes that cannot be performed physically, e.g., from 
the inside of an object bundle or in narrow spaces between objects, are filtered out 
based on the masks returned by R-CNN. In the figure, for example, samples between the 
two object bundles are deleted. A sampled push is defined as $p = (x, y, \theta) \in 
SE(2)$ where $x$, $y$ are the start location of the push and $\theta$ indicates the 
horizontal push direction. The push distance is fixed.

\textbf{Input to push prediction}. The initial scene image $I$, scene object masks 
and centers $(m_i, c_i)$, and a sampled push action $p = (x, y,\theta)$ are the main 
inputs to the push prediction module. To reduce redundancy, we transform all inputs 
using a 2D homogeneous transformation matrix $T$ such that $pT = (40, 112, 0)$. The position of the push is normalized for easier learning such that a push will always go from left to right in the middle of the left side of the workspace.
From here, it is understood that all inputs are with respect to this updated 
coordinate frame defined by $T$, i.e., $p \leftarrow pT, c_i \leftarrow c_i T$, 
and so on. 
Apart from the inputs mentioned so far, we also generate: (1) one binary 
\emph{push action} image $M_p$ with all pixels black except in a small square with 
top-left corner $(40, 100)$ and bottom-right corner $(65, 124)$ which is the finger movement space, (2) one $224\times 
224$ binary image $M_I$ with the foreground of $I$ set to white, and (3) one $60 \times 
60$ binary mask image $M_i$ for each object mask $m_i$, centered at $c_i$. 
Despite being constant relative to the image transformed by $T$, push image $M_p$ is used 
as an input because we noticed from our experiments that it helps the network focus 
more on the pushing area. 

\textbf{Push prediction.} 
With global (binary images $M_p, M_I$) and local (mask image $M_i$ and the center $c_i$ 
of each object) information, \dipn proceeds to predict objects' transformations. 
To start, a Multi-Layer Perceptron (MLP) and a ResNet~\cite{DBLP:journals/corr/LinDGHHB16} 
(with no pre-training) are  used to encode the push action and the global information, 
respectively: 
\[e_p = \text{MLP}_\text{1}(p),\quad e_{AB} = \text{ResNet}_\text{1}(M_p,M_I).\]
A similar procedure is applied to individual objects. 
For each object $o_i$, its center $c_i$ and mask image $M_i$ are encoded using ResNet 
(again, with no pre-training) and MLP as: 
\[e_i = (\text{ResNet}_\text{2}(M_i),  \text{MLP}_\text{2}(c_i)).\]
Adopting the design philosophy from~\cite{watters2017visual}, the encoded information 
is then passed to 
a {\em direct transformation} (DT) MLP module 
(blocks in Fig.~\ref{fig:push-prediction-flowchart} with orange background) 
and multiple {\em interactive transformation} (IT) MLP modules 
(blocks in Fig.~\ref{fig:push-prediction-flowchart} with cyan background): 
\[\forall 1\leq i \leq n:\ \text{DT}_i = \text{MLP}_\text{3}(e_p, e_i), \]
\[\forall 1\leq i, j \leq n, j \neq i:\ \text{IT}_{ij} = \text{MLP}_\text{4}(e_p, e_i, e_j).\]
Here, the direct transformation modules capture the effect of the robot {\em directly} 
touching the objects (if any), while the interactive transformation modules consider 
collision between an object and every other object (if any).
Then, all aforementioned encoding is put together to a decoding MLP to derive the output 
2D transformation for each object $o_i$ in the push action's frame: $\forall 1 {\leq} i {\leq} n$, 
\[(\hat x_i, \hat y_i, \hat \theta_i) = 
\text{MLP}_\text{5}(e_{AB}, \text{DT}_i + \sum_{1\leq j \leq n, j \neq i} \text{IT}_{ij}),\]
which can be mapped back to the original coordinate frame via $T^{-1}$. This yields  
predicted poses of objects. 
From these, an ``imagined'' push prediction image is readily generated. 

In our implementation, both ResNet\textsubscript{1} and ResNet\textsubscript{2} are 
ResNet-50. MLP\textsubscript{1} and MLP\textsubscript{2}, encoding the push action 
and single object position, both have two (hidden) layers with sizes 8 and 16. 
MLP\textsubscript{3}, connecting encoded and direct transformations, has two layers 
of a uniform size 128. MLP\textsubscript{4}, connecting encoded and interactive 
transformations, has three layers with a size of 128 each. 
The final decoder MLP\textsubscript{5} has five layers with sizes [256, 64, 32, 16, 3]. 
The number of objects $n$ is not fixed and generally varies across scenes. The network 
can nevertheless handle varying and unbounded numbers of objects because the same 
networks MLP\textsubscript{2}, MLP\textsubscript{3}, MLP\textsubscript{4}, MLP\textsubscript{5}, 
and ResNet\textsubscript{2} are used for all the objects.

\textbf{Training.}
For training in simulation and for real experiments, objects are randomly dropped 
onto the workspace. The robot then executes random pushes to collect training data.
\emph{SmoothL1Loss} (Huber Loss) is used as the loss function. 
Given each object's true post-push transformation $(x_i, y_i, \theta_i)$ and the 
predicted $(\hat x_i, \hat y_i, \hat \theta_i)$, the loss is computed as the sum of 
coordinate-wise SmoothL1Loss between the two.
\dipn performs well on unseen objects and can be completely trained in simulation 
and transferred to the real-world (Sim-to-Real). It is also robust with respect to 
changes in objects' physical properties, e.g., variations in mass and friction coefficients. 

\subsection{The Grasp Network (\gn)}
We briefly describe \gn, which shares a similar architecture to the DQN used 
in~\cite{zeng2018learning}. Given an observed image and candidate grasp actions as 
described in Section~\ref{sec:preliminaries}, \gn finds the optimal policy 
for maximizing a single-step grasp reward, defined as $1$ for a successful grasp 
and $0$ otherwise. 
\gn focuses its \emph{attention} on local regions that are relevant to each single 
grasp and uses image-based self-supervised pre-training to achieve a good
initialization of network parameters. 

\begin{figure}[ht!]
    \centering
    \vspace*{14pt}
    \begin{overpic}[width=\linewidth]{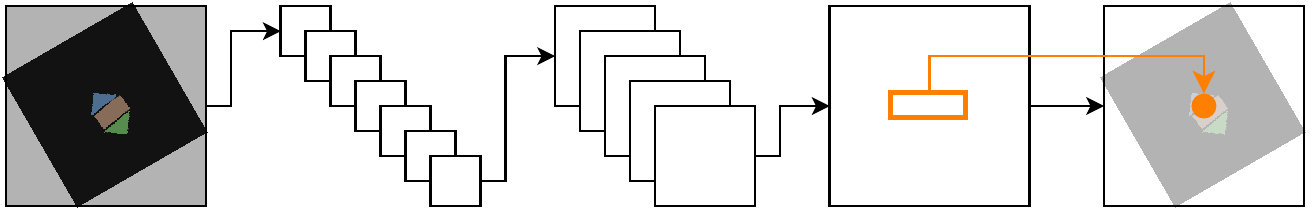}
        \scriptsize
        \put (8 , 18) {\makebox(0,0){\textcolor{NavyBlue}{$320{\times}320{\times}4$}}}
        \put (29, 18) {\makebox(0,0){\textcolor{RubineRed}{$256$}-\textcolor{NavyBlue}{$80{\times}80$}}}
        \put (50, 18) {\makebox(0,0){\textcolor{RubineRed}{$64$}-\textcolor{NavyBlue}{$160{\times}160$}}}
        \put (70, 18) {\makebox(0,0){\textcolor{RubineRed}{$1$}-\textcolor{NavyBlue}{$320{\times}320$}}}
        \put (91.5, 18) {\makebox(0,0){\textcolor{NavyBlue}{$320{\times}320{\times}1$}}}
        \put (18, -1.5) {\makebox(0,0){ResNet-50}}
        \put (18, -4) {\makebox(0,0){FPN P2}}
        \put (37, -1.5) {\makebox(0,0){Conv \textcolor{RubineRed}{128}, \textcolor{ForestGreen}{3$\times$3}}}
        \put (37, -6.5) {\makebox(0,0){\centering ReLU}}
        \put (37, -4) {\makebox(0,0){Batchnorm2d(\textcolor{RubineRed}{128})}}
        \put (37, -9) {\makebox(0,0){Conv \textcolor{RubineRed}{128}, \textcolor{ForestGreen}{3$\times$3}}}
        \put (37, -11.5) {\makebox(0,0){Batchnorm2d(\textcolor{RubineRed}{128})}}
        \put (37, -14) {\makebox(0,0){\centering ReLU}}
        \put (37, -16.5) {\makebox(0,0){\centering Interpolate 2$\times$}}
        \put (61, -1.5) {\makebox(0,0){Conv \textcolor{RubineRed}{32}, \textcolor{ForestGreen}{3$\times$3}}}
        \put (61, -4) {\makebox(0,0){Batchnorm2d(\textcolor{RubineRed}{32})}}
        \put (61, -6.5) {\makebox(0,0){\centering ReLU}}
        \put (61, -9) {\makebox(0,0){\centering Dropout(0.1)}}
        \put (61, -11.5) {\makebox(0,0){Conv \textcolor{RubineRed}{1}, \textcolor{ForestGreen}{3$\times$3}}}
        \put (61, -14) {\makebox(0,0){\centering Interpolate 2$\times$}}
        \put (83, -1.5) {\makebox(0,0){Conv \textcolor{RubineRed}{1}, \textcolor{ForestGreen}{11$\times$57}}}
        \put (83, -4) {\makebox(0,0){ReLU}}
        \put (83, -6.5) {\makebox(0,0){Conv \textcolor{RubineRed}{1}, \textcolor{ForestGreen}{11$\times$57}}}
    \end{overpic}
    \vspace*{35pt}
    \caption{\label{fig:grasp-network}
    Architecture of \gn.
    Pink, blue, and green text are used for channel count, image size, and kernel 
    size, respectively. 
    }
\end{figure}
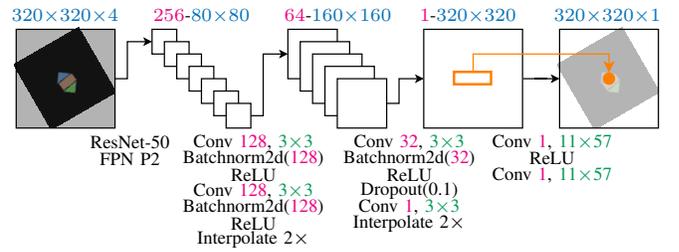

The proposed modified network's architecture is illustrated in Fig.~\ref{fig:grasp-network}. 
It takes an input image and outputs a score for each candidate grasp centered at 
each pixel. The input image is rotated to align it with the end-effector frame 
(see the left image in Fig.~\ref{fig:grasp-network}).  
 ResNet-50 FPN~\cite{DBLP:journals/corr/LinDGHHB16} is used as the backbone; 
we replace the last layer with our own customized head structure shown in Fig.~\ref{fig:grasp-network}. 
We observe that our structure leads to faster training and inference 
time without loss of accuracy. 
Given that the network computes pixel-wise values 
in favor of a local grasp region, at the end of the network, we place two 
convolutional layers with kernel size $11 \times 57$, which was determined based 
on the clearance of the gripper.

In training \gn, image-based pre-training~\cite{yen2020learning} 
was employed. The pre-training process treats pixel-wise grasping as a vision task 
to obtain a quality network initialization. The process automatically labels with $0$ 
or $1$ all the pixels in a small set of arbitrary images, depending on whether grasps 
centered at each pixel would lead to a finger collision with an object, based only on 
color/depth and without actually simulating or executing the grasps physically. 
The pre-training data set does not include objects used for testing. 

These design choices render \gn sample efficient and fairly accurate, as shown in 
Section~\ref{sec:evaluation}.

\subsection{The Complete Algorithmic Pipeline}

The training process of \dipngn is outlined in Alg.~\ref{alg:train}. In 
line~\ref{alg:train-pre}, an image data set is collected for training Mask R-CNN 
(i.e., for push prediction segmentation) and initializing (i.e., pre-training) \gn. 
Note that training data for Mask R-CNN and pre-training data for GN are essentially 
free, with no physics involved. After pre-training, the training process for push 
(line~\ref{alg:train-push}) and grasp (line~\ref{alg:train-grasp}) predictions can 
be executed on \pag scenes in any order. 

\newcommand\mycommfont[1]{\footnotesize\normalfont{#1}}
\SetCommentSty{mycommfont}
\begin{algorithm}
    \small
    \DontPrintSemicolon
    \KwOut{trained \dipngn.}
    \gn, Mask R-CNN${\gets}$GetImageDataSetAndPre-Train\,$()$\label{alg:train-pre}\;
    \dipn $\gets$ TrainOn{\pag}PushOnly\,$($Mask R-CNN$)$ \label{alg:train-push}\;
    \gn $\gets$ TrainOn{\pag}GraspOnly\,$($GN$)$ \label{alg:train-grasp}\; 
    \caption{\label{alg:train}
    Training \dipngn
    }
\end{algorithm}

The high-level workflow of our framework on \pag is described in Alg.~\ref{alg:evaluation}. 
When working on an instance, at every decision-making step $t$, an image $M_t$ is first obtained 
(line~\ref{alg:evaluation-input}). 
Then, the image $M_t$, along with sampled push actions $A^{\text{push}}$, are sent to 
the trained \dipn to generate predicted synthetic images $\hat{M}_{t+1}$ after each imagined 
push $a$ (line~\ref{alg:evaluation-sample}-\ref{alg:evaluation-dipn}). 
With $A^{\text{grasp}}$ denoting the set of all grasp actions, 
their discounted average reward on the predicted next image $\hat{M}_{t+1}$ is then 
compared with the average of grasping rewards in the current image 
(line~\ref{alg:evaluation-compare}): recall that \gn takes an image and a grasp action as 
input, and outputs a scalar grasp reward value. If there exists a push action with a higher 
expected average grasping reward in the predicted next image, the best push action is then 
selected and executed (line~\ref{alg:evaluation-push}); otherwise, the best grasp action 
is selected and executed (line~\ref{alg:evaluation-grasp}). %
Because it is desirable to have a single push action that simultaneously renders multiple 
objects graspable, the average grasp reward is used instead of only the maximum. 

\begin{algorithm}
    \small
    \DontPrintSemicolon
    \KwIn{trained \gn and \dipn, discount factor $\gamma$}
    \While{\normalfont there are objects in workspace}{
        $A^{\text{push}} \gets \varnothing$, $M_t \gets$ GetImage\,$()$; \label{alg:evaluation-input} \; 
        \For{\normalfont $a$ in SamplePushActions\,$(M_t)$}{ \label{alg:evaluation-sample}
            $A^{\text{push}} \gets A^{\text{push}} \cup \{a\}$; $\hat{M}_{t+1} \gets$ \dipn$(M_t, a)$; \label{alg:evaluation-dipn}\;
            $Q(M_t, a) = \frac{\gamma}{|A^{\text{grasp}}|} \sum_{a'\in A^{\text{grasp}}} \text{\gn}(\hat{M}_{t+1}, a')$;
        }
        \If{\normalfont $\max_{a \in A^{\text{push}}} Q(M_t, a) > \frac{1}{|A^{\text{grasp}}|} \sum\limits_{a'\in A^{\text{grasp}}} \text{\gn}(M_t, a')$}{ \label{alg:evaluation-compare}
            Execute $\argmax_{a \in A^{\text{push}}} Q(M_t, a)$;
            \label{alg:evaluation-push}
        }
        \lElse{
            Execute $\argmax_{ a\in A^{\text{grasp}}  }\text{\gn}(M_t, a)$; \label{alg:evaluation-grasp}
        }
    }
    \caption{\label{alg:evaluation}
    Executing \dipngn}
\end{algorithm}

The framework contains two hyperparameters. 
The first one, $\gamma$, is the discount factor of the Markov Decision Process. 
For a push action to be selected, the estimated discounted grasp reward after a 
push must be larger than grasping without a push, since the push and then grasp 
takes two actions. In our implementation, we set $\gamma$ to be $0.9$. 
The other {\em optional} hyperparameter is used for accelerating inference: if 
the maximum grasp reward is higher than a threshold, we directly execute the 
grasp action without calling the push prediction. This hyperparameter requires 
tuning. In our implementation, the threshold value is set to be $0.7$. Note that 
the maximum reward for a single grasp is $1$. 


\section{Experimental Evaluation}\label{sec:evaluation}

We first evaluate \gn and \dipn separately and then compare the full system's 
performance with the state-of-the-art model-free RL technique presented 
in~\cite{zeng2018learning}, which is the closest work to ours. 
Apart from evaluating our approach on a real robotic system, we also did 
extensive evaluations in the CoppeliaSim~\cite{rohmer2013v} simulator. 
We use an Nvidia GeForce RTX 2080 Ti graphics card to train and test the
algorithms. 
All simulation experiments are repeated $30$ times; all real experiments 
are repeated for $5$ times to get the mean metrics.

\subsection{Deep Interaction Prediction Network (\dipn)}
To evaluate how accurately \dipn can predict the next image after a push action,  
\dipn is first trained on randomly generated \pag instances in simulation. 
At the start of each episode, randomly generated objects with random colors 
and shapes are randomly dropped from mid-air to construct the scene. 
Up to $7$ objects are generated per scene.  
We calculate the prediction accuracy by measuring the {\it Intersection-over-Union} 
(IoU) between a predicted image and the corresponding ground-truth after pushing.
The IoU calculation is performed at the object level and then averaged.

\begin{figure}[ht!]
    \centering
    \includegraphics[width = \linewidth]{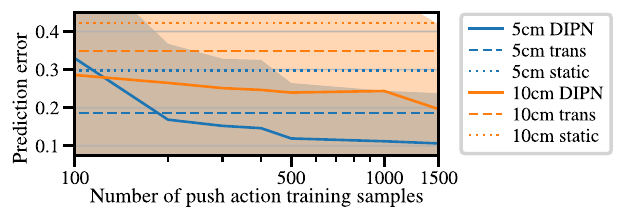}
    \caption{\label{fig:result-push-prediction}
    \dipn learning curve with standard deviation visualized as colored areas. 
    The $x$-axis is the number of pushes for training \dipn.
    The $y$-axis is the prediction error: $1-\text{IoU}$.
    The dotted and dashed lines are baselines.}
\end{figure}

\dipn is compared with two baselines: 
the first one, called {\em static}, assumes that all objects stay still. 
The second one, called {\em trans}, always assumes that only the pushed object
moves, and that it moves exactly by the push distance along the push direction. 
Both baselines are engineered methods that do not require training. 
The push prediction errors ($1-\text{IoU}$) are illustrated in 
Fig.~\ref{fig:result-push-prediction} as learning curves in simulation. 
Mask R-CNN is trained (i.e., Alg.~\ref{alg:train}, line~\ref{alg:train-pre}) 
using an additional $100$ images, which is why Fig.~\ref{fig:result-push-prediction} 
starts from $100$. 
We observe, for different push distances, that \dipn outperforms the baselines 
with a large margin after sufficient training. 
After convergence, the prediction error for \dipn is less than $0.1$ for a $5$cm 
push, which indicates that the predicted pose of an object overlaps $90\%+$ with 
the ground truth. 
As expected, \dipn is more accurate and more sample efficient with a shorter 
push distance. On the other hand, longer push distances generally result in better
overall performance for \pag challenges even though push predictions become 
less accurate, since larger actions are more effective in terms of changing the scene.

\begin{figure}[ht!]
    \centering
    \includegraphics[trim={60 60 120 120}, clip, width=0.24\linewidth]{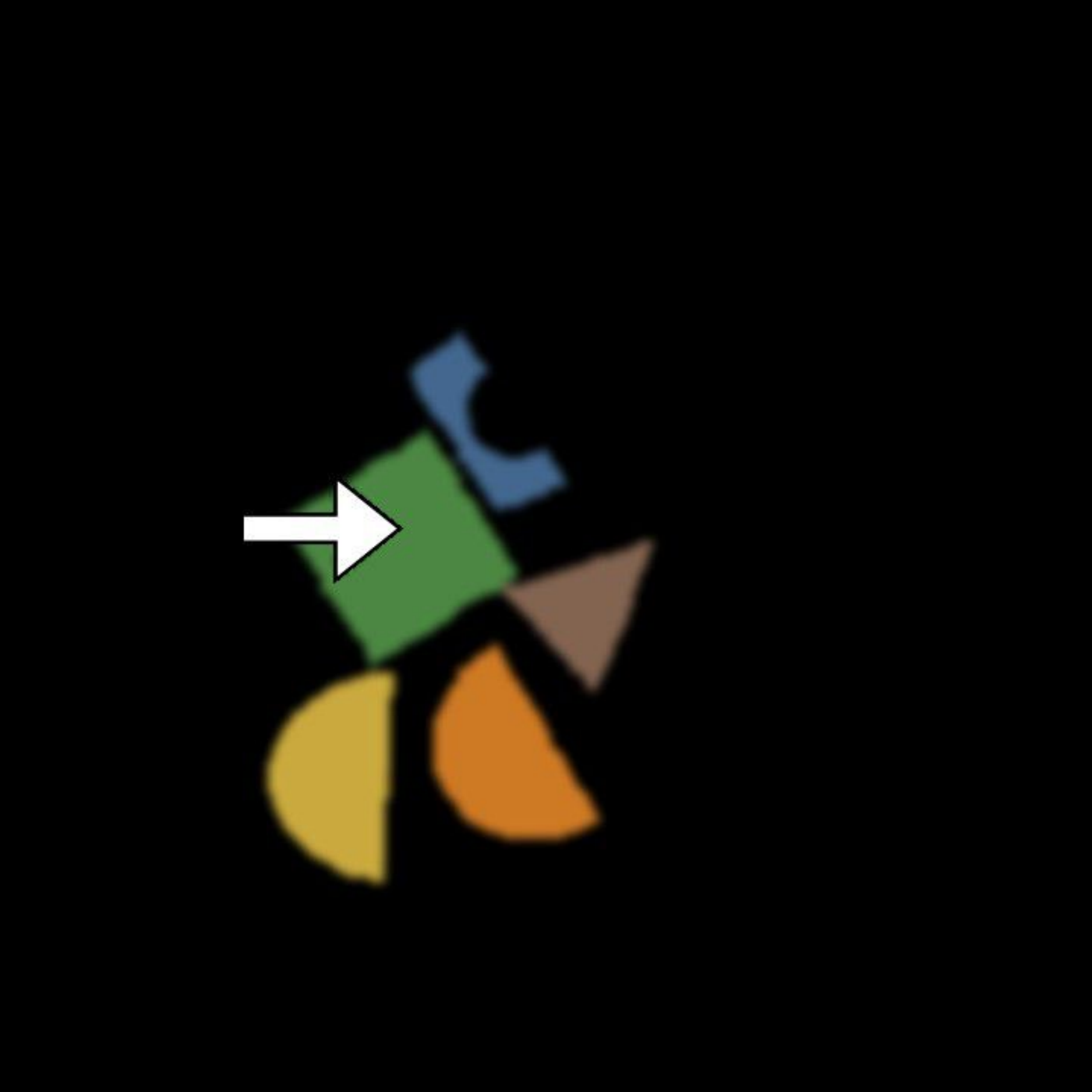}
    \hfill
    \includegraphics[trim={60 60 120 120}, clip, width=0.24\linewidth]{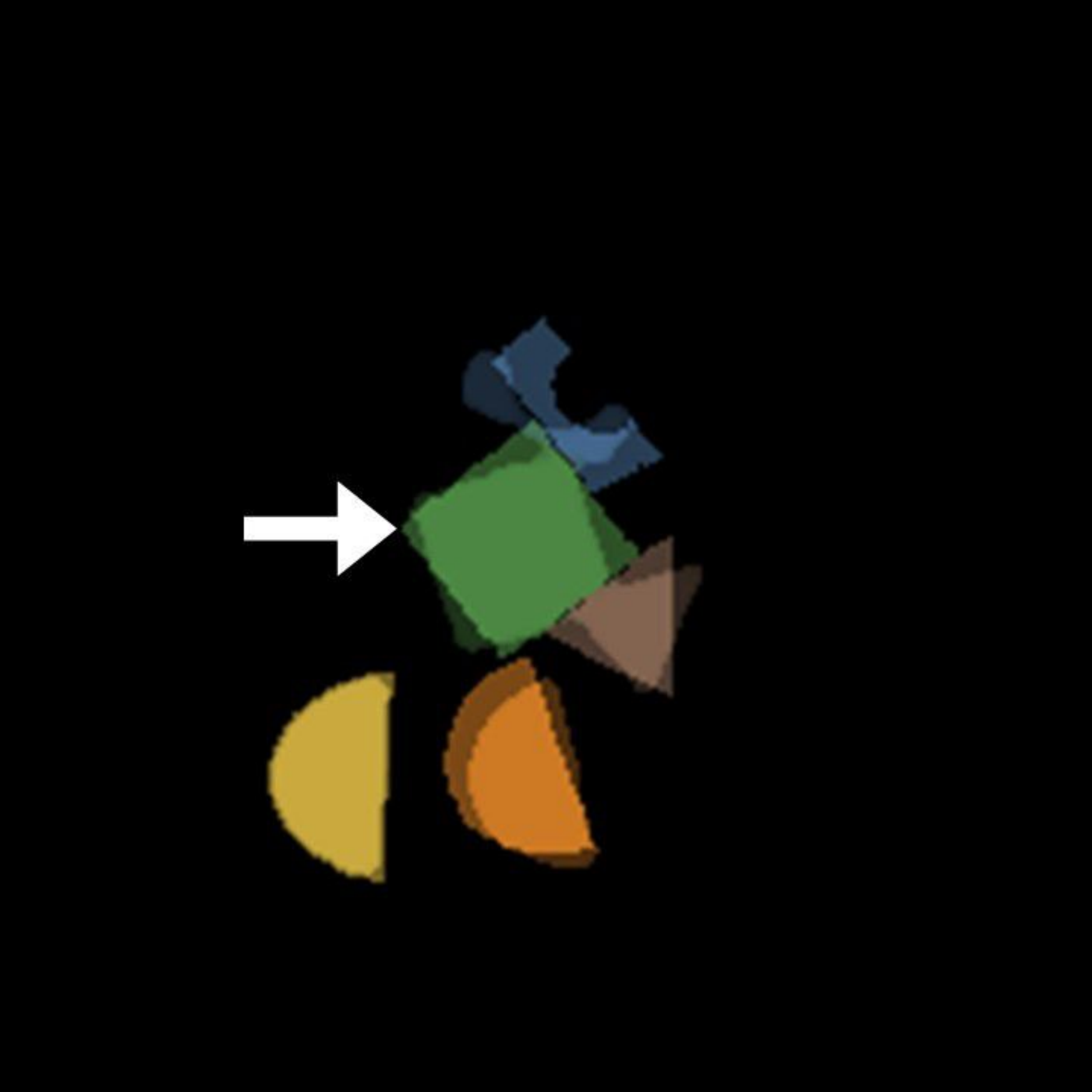}
    \hfill
    \includegraphics[trim={85 120 95 60}, clip, width=0.24\linewidth]{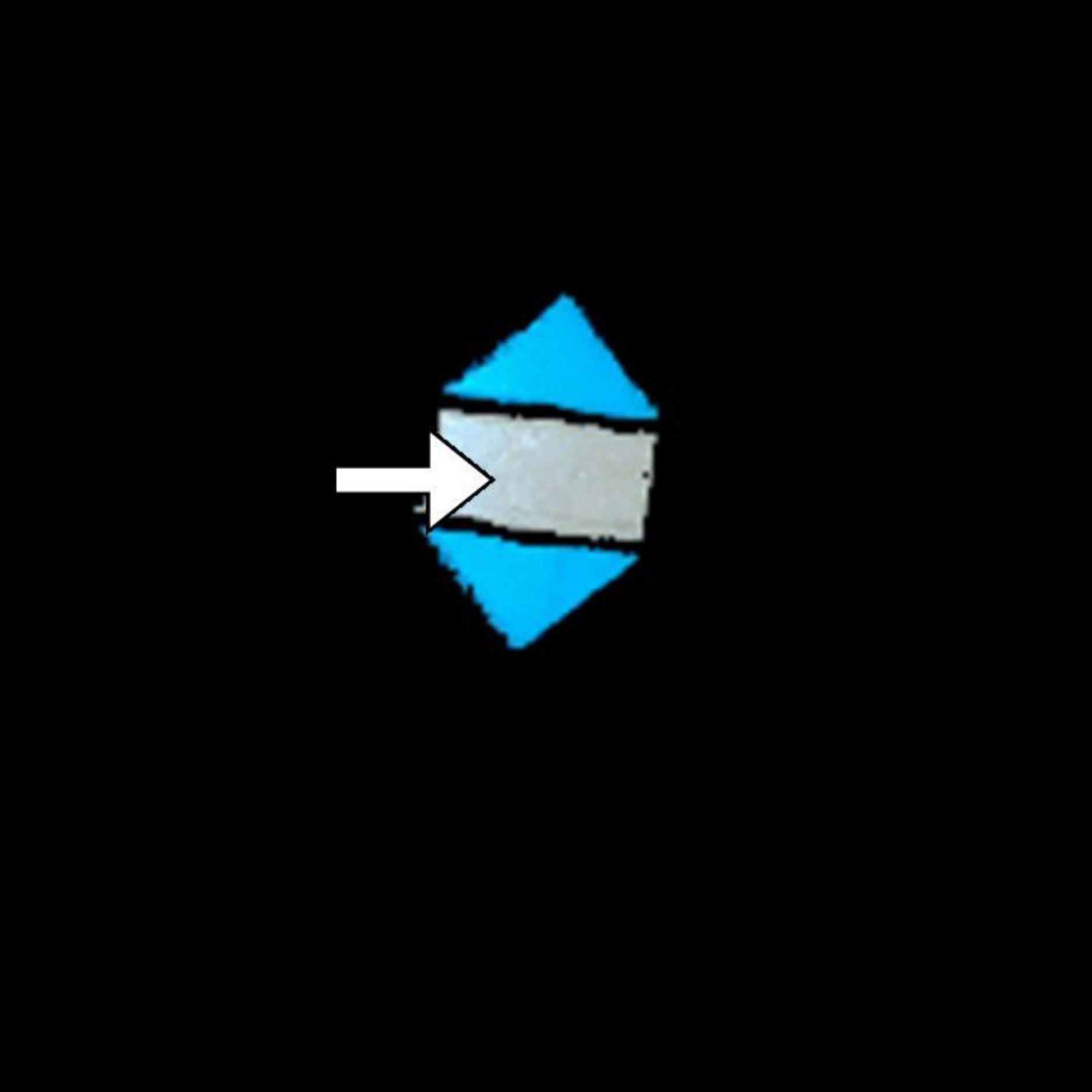}
    \hfill
    \includegraphics[trim={85 120 95 60}, clip, width=0.24\linewidth]{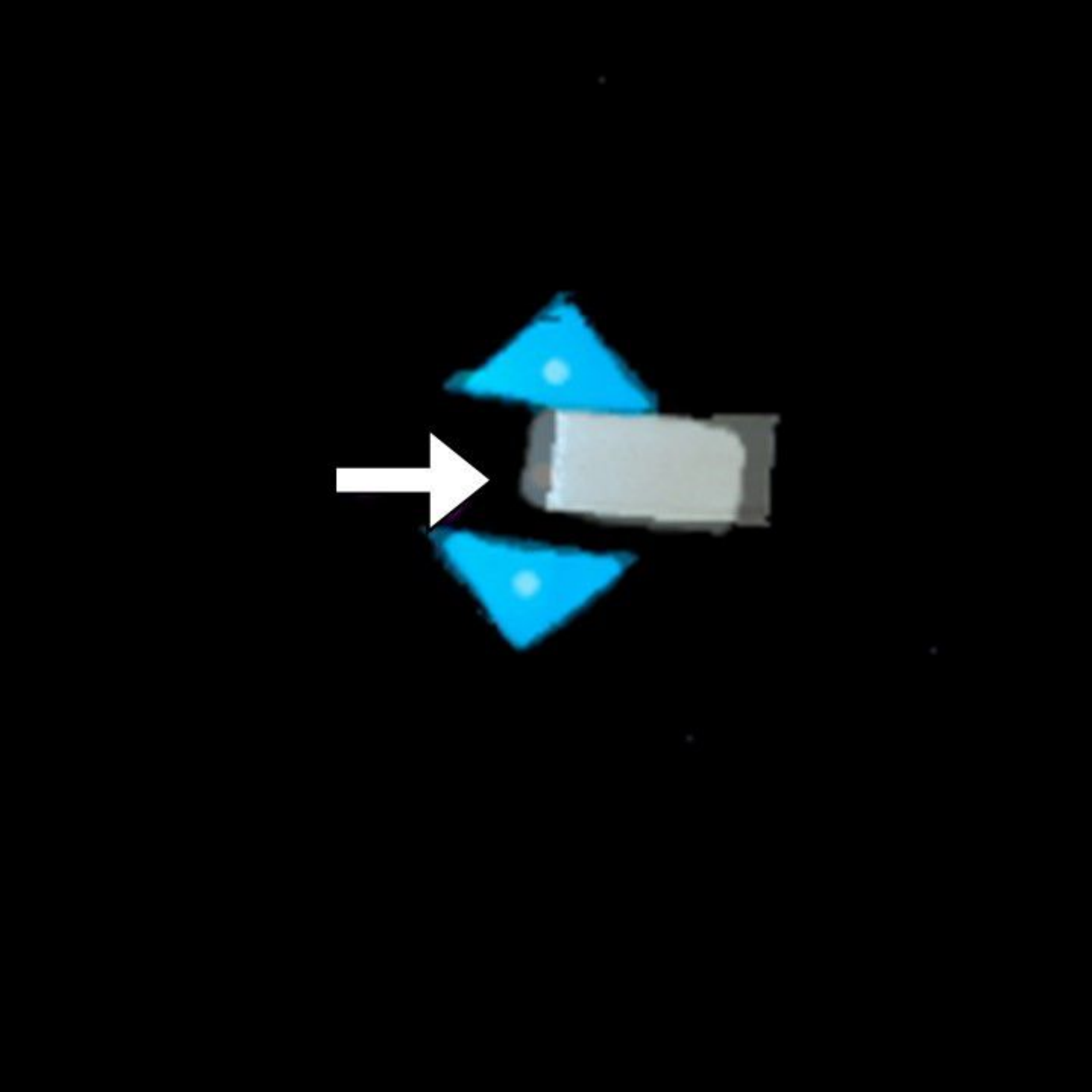}
    \caption{\label{fig:push-prediction-result} 
    Typical \dipn results. 
    The figures from left to right are: 
    original and predicted images in simulation, and original and predicted images 
    in a real experiment. The ground truth images after a push are overlaid on the
    predicted images with transparency.     The arrows visualize the push actions. 
    }
\end{figure}

Fig.~\ref{fig:push-prediction-result} shows typical predictions by \dipn. 
The network is learned in simulation with randomly shaped and colored objects, and 
directly transferred to the real system. We observe that \dipn can accurately predict 
the state after a push, with good accuracy on object orientation and translation. 

\begin{figure*}[ht!]
    \centering
    \includegraphics[width = 0.189 \linewidth, trim = {450, 200, 450, 320}, clip]{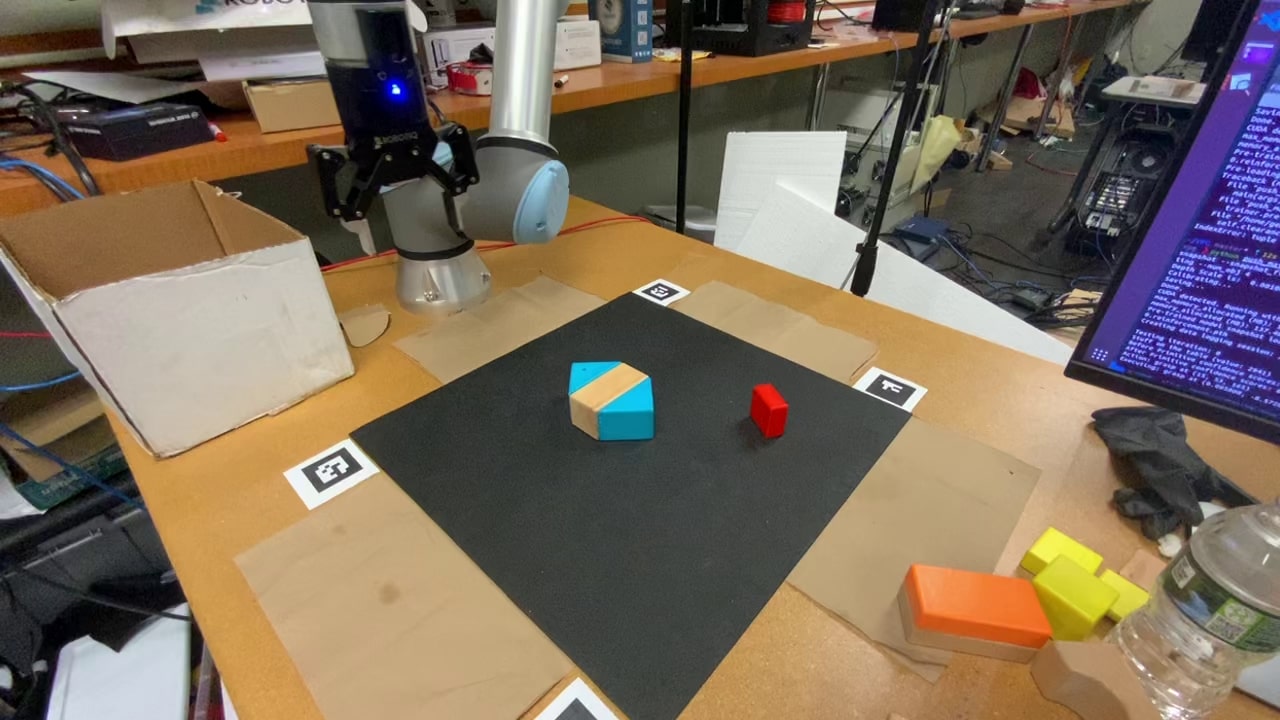} \hfill
    \includegraphics[width = 0.189 \linewidth, trim = {450, 200, 450, 320}, clip]{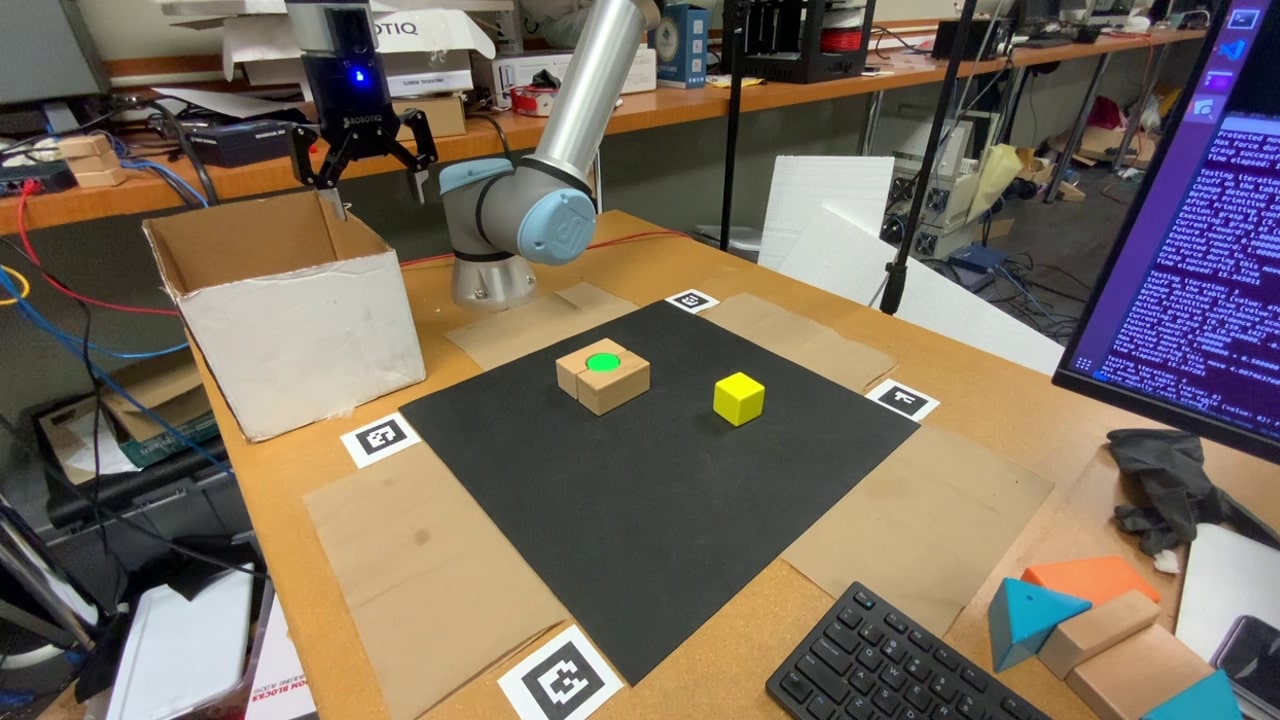} \hfill
    \includegraphics[width = 0.189 \linewidth, trim = {450, 200, 450, 320}, clip]{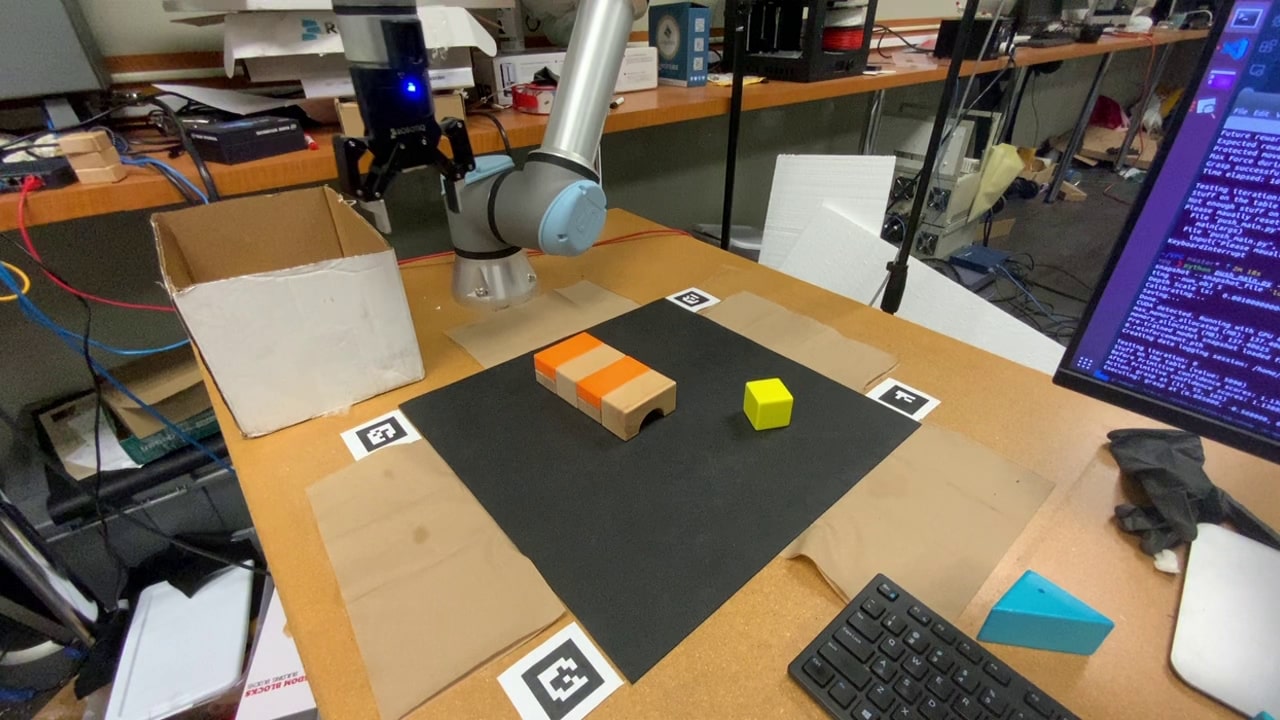} \hfill
    \includegraphics[width = 0.189 \linewidth, trim = {450, 200, 450, 320}, clip]{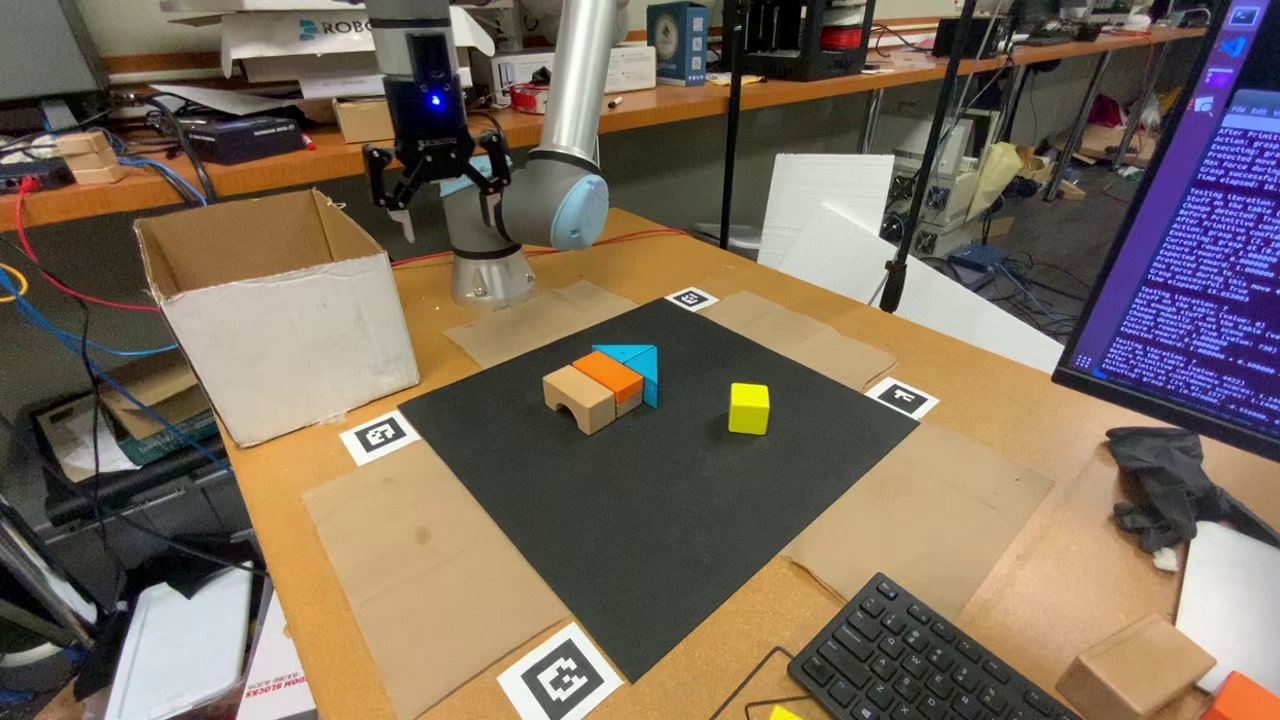} \hfill
    \includegraphics[width = 0.189 \linewidth, trim = {450, 200, 450, 320}, clip]{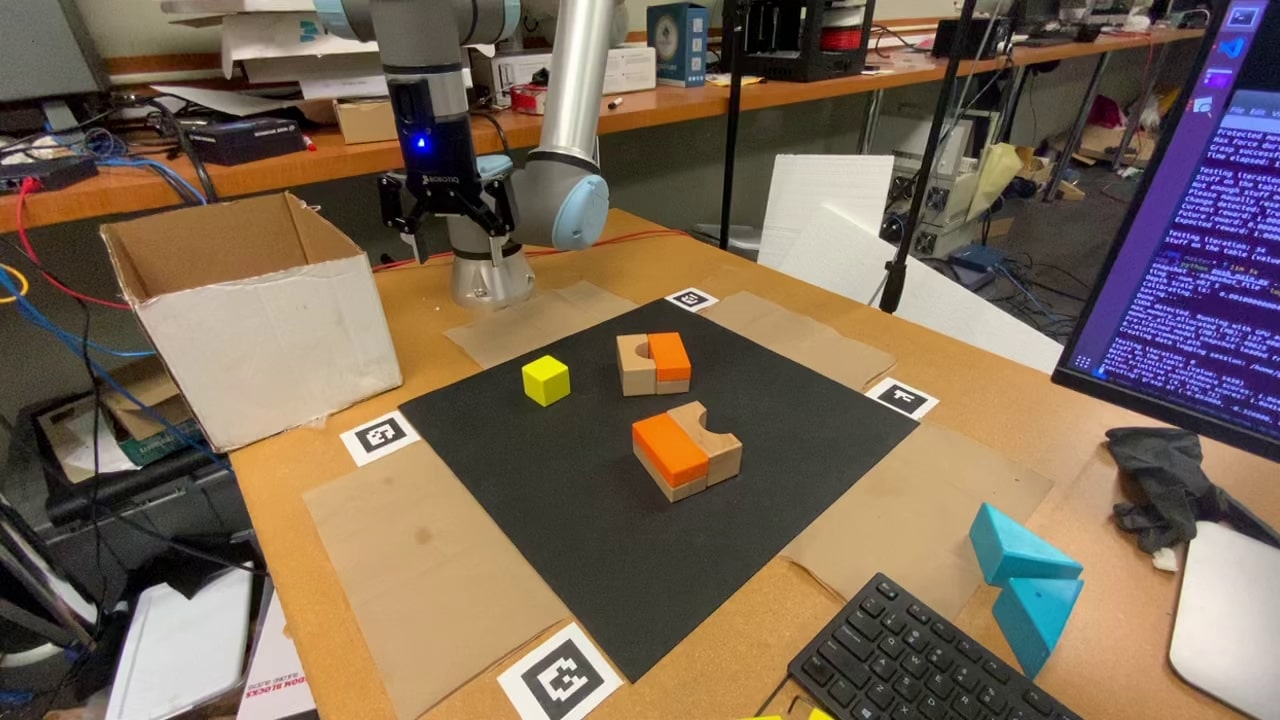}
    
    \vspace*{5pt}
    
    \includegraphics[width = 0.189 \linewidth, trim = {450, 200, 450, 320}, clip]{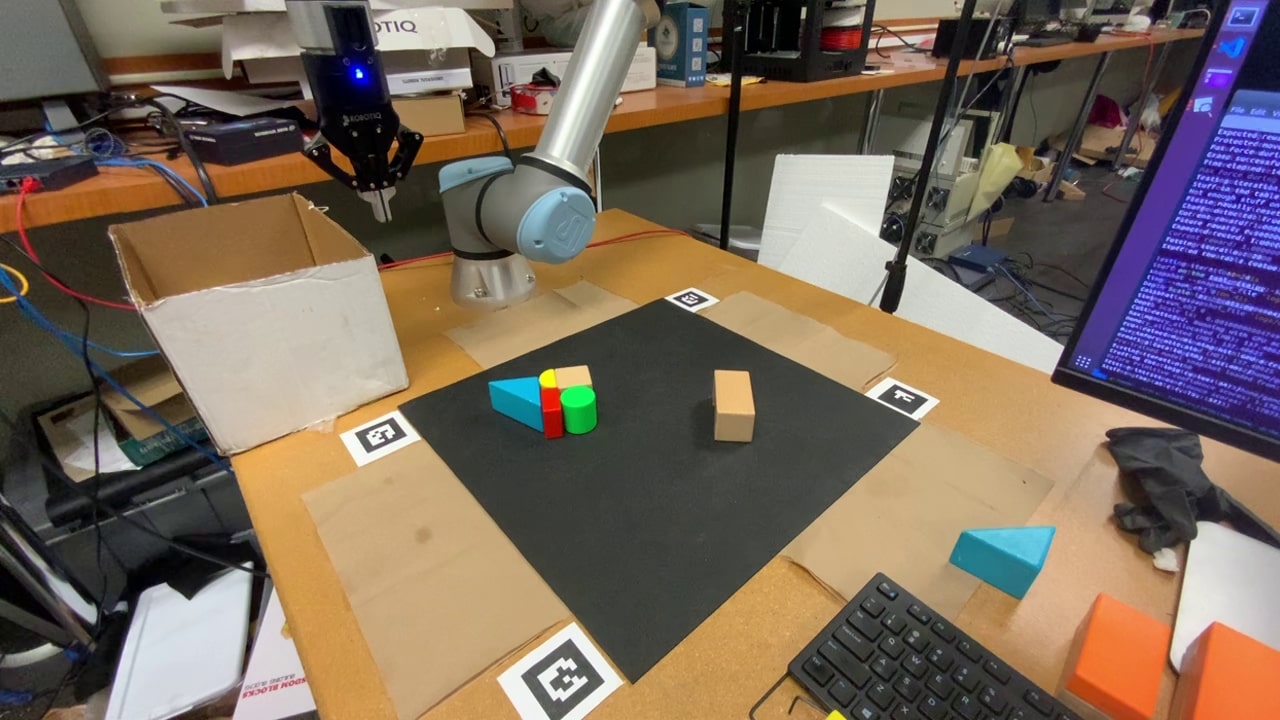} \hfill
    \includegraphics[width = 0.189 \linewidth, trim = {450, 200, 450, 320}, clip]{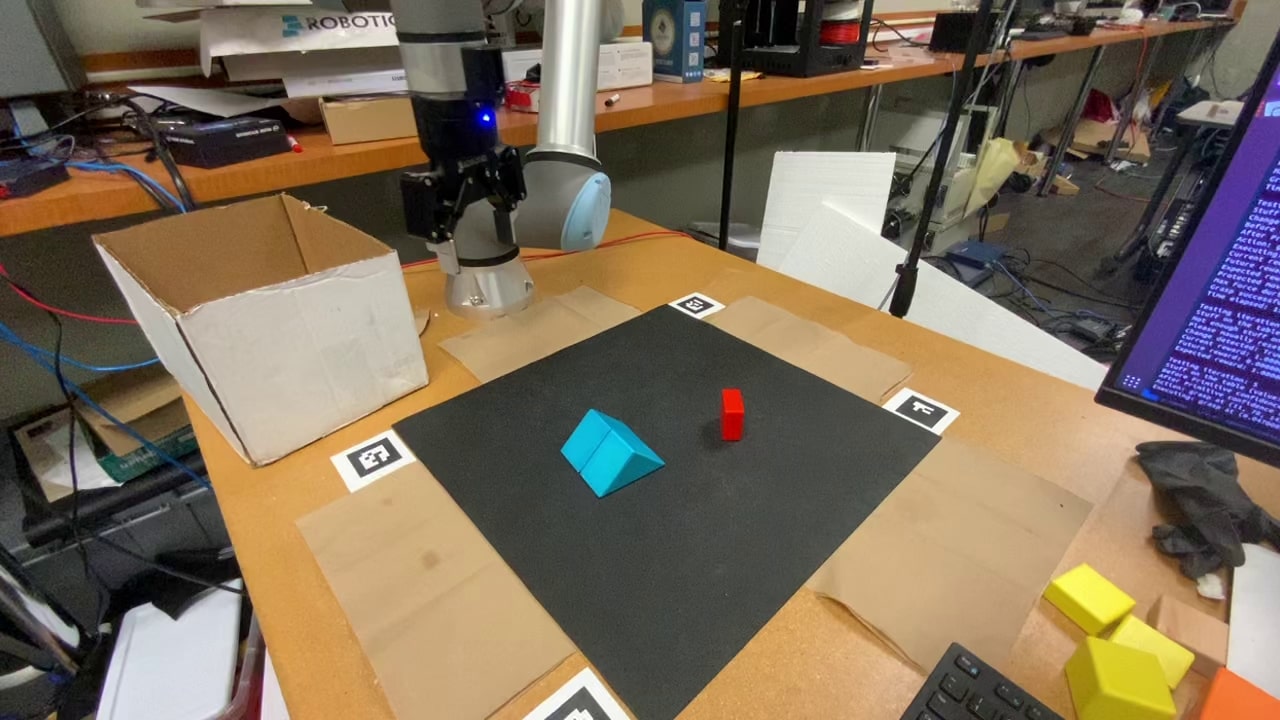} \hfill
    \includegraphics[width = 0.189 \linewidth, trim = {450, 200, 450, 320}, clip]{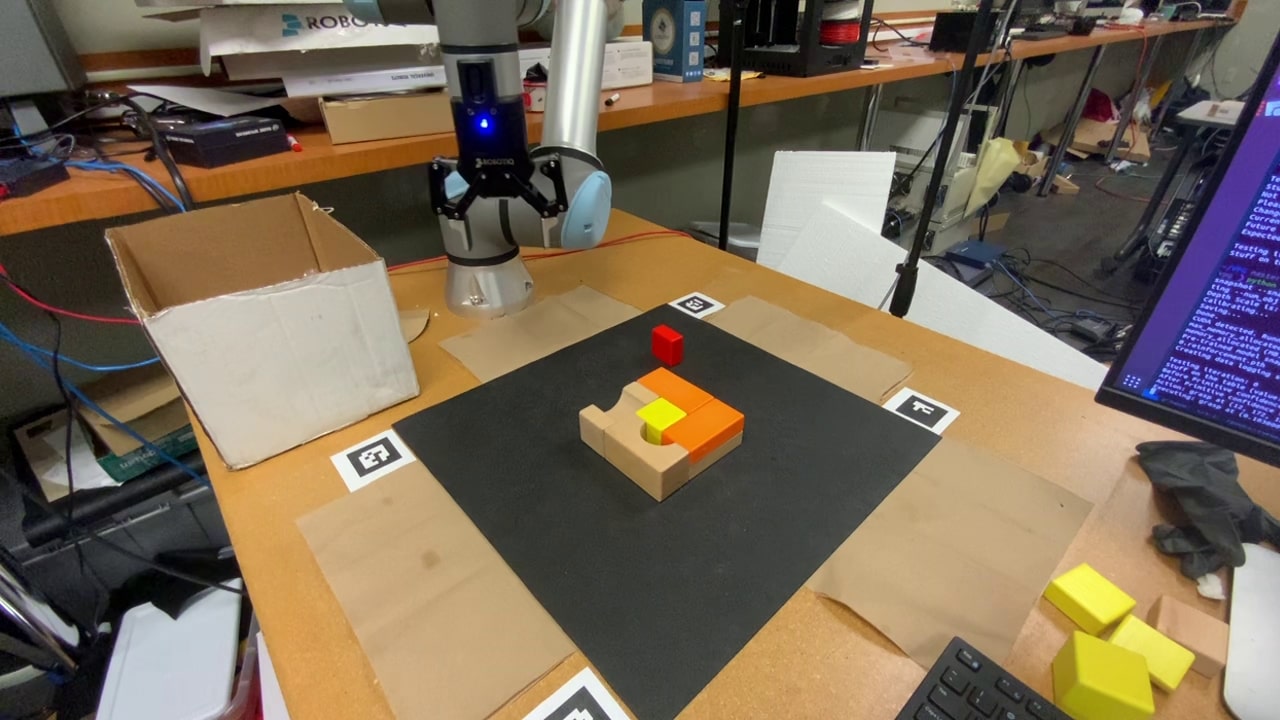} \hfill
    \includegraphics[width = 0.189 \linewidth, trim = {450, 200, 450, 320}, clip]{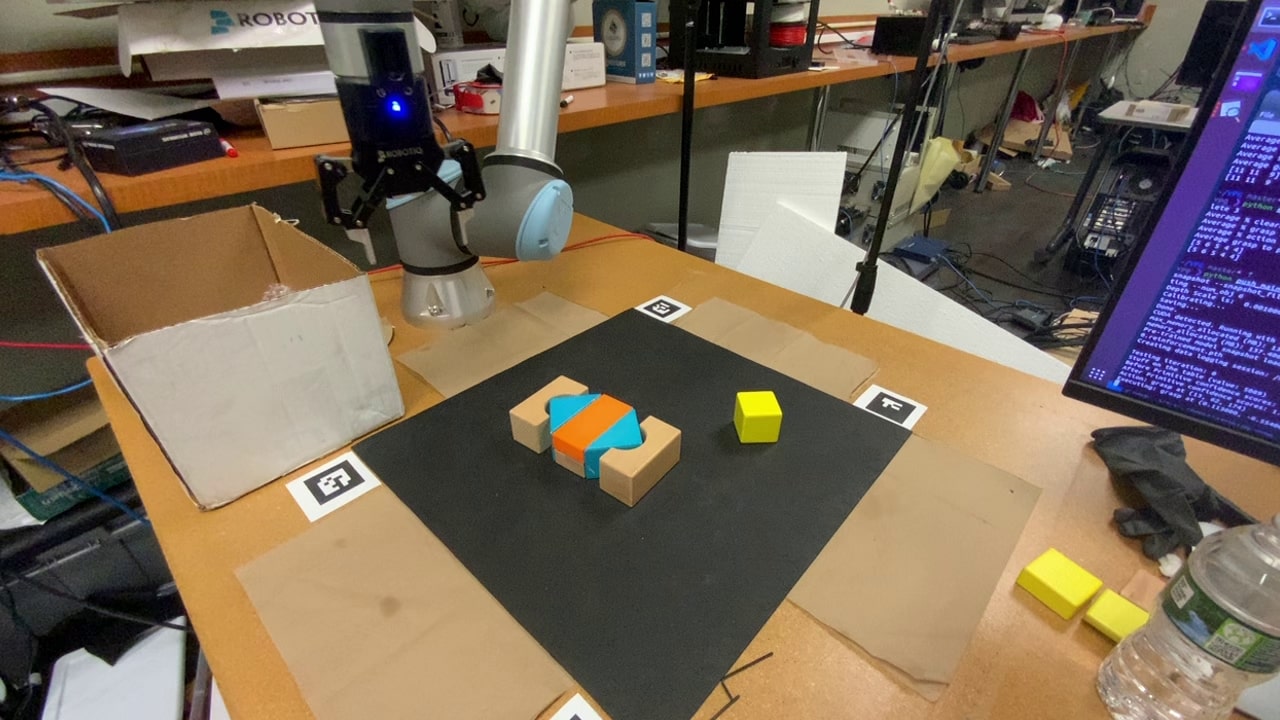} \hfill
    \includegraphics[width = 0.189 \linewidth, trim = {450, 200, 450, 320}, clip]{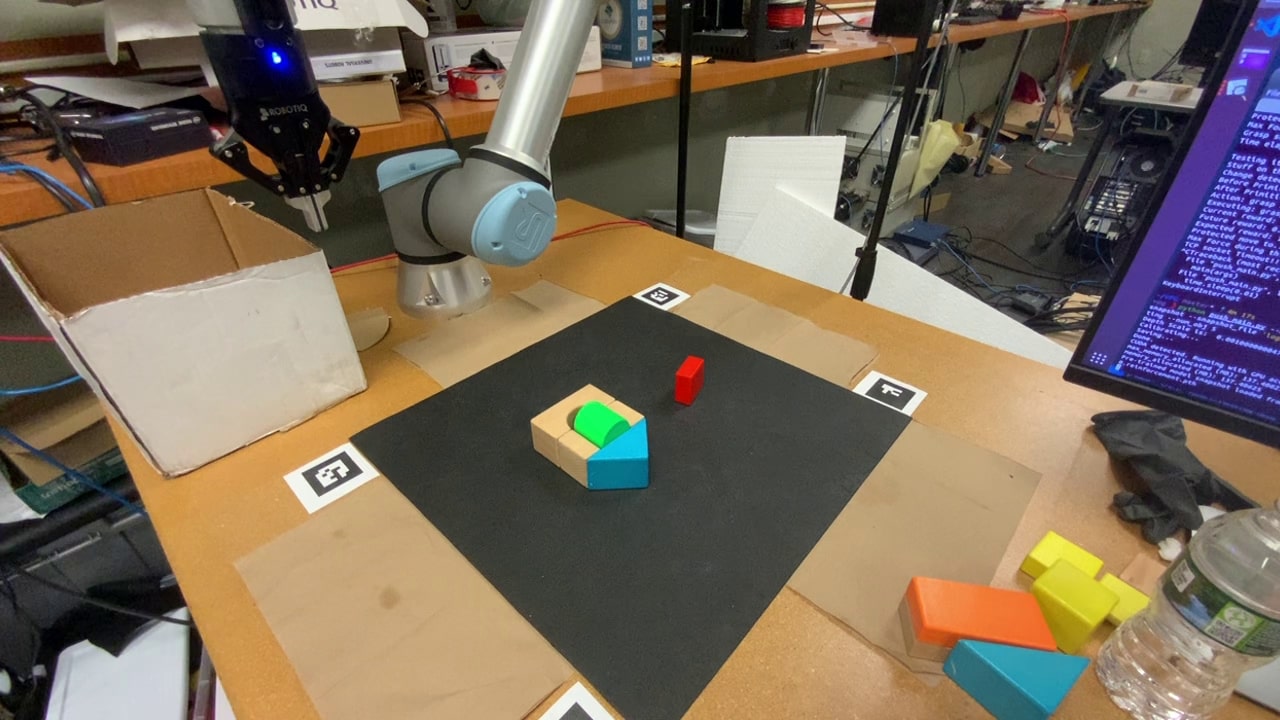}
    \caption{\label{fig:evaluation-hard-instances}
    Manually generated hard instances largely similar to the ones in \cite{zeng2018learning}. 
    The cases are used in both simulation and real experiment.
    }
\end{figure*}

\subsection{Grasp Network (\gn)}
We train and evaluate \gn as a standalone module and compare it with the state-of-the-art 
DQN-based method known as Visual Pushing and Grasping (VPG)~\cite{zeng2018learning}, 
which learns both grasp and push at the same time. 
Since \gn is only trained on grasp actions, and for a fair comparison, we also tested a
third method that learns both grasp and push actions: this method, denoted by DQN+\gn,
uses \gn for learning grasp actions and the DQN structure of \cite{zeng2018learning} to 
learn push actions. 
The algorithms are compared using the grasp success rate metric, i.e., {\em the number 
of objects removed} divided by {\em the total number of grasps}.
We train all algorithms directly on randomly generated \pag instances with $10$ objects.

\begin{figure}[ht!]
    \centering
    \includegraphics[width =  \linewidth]{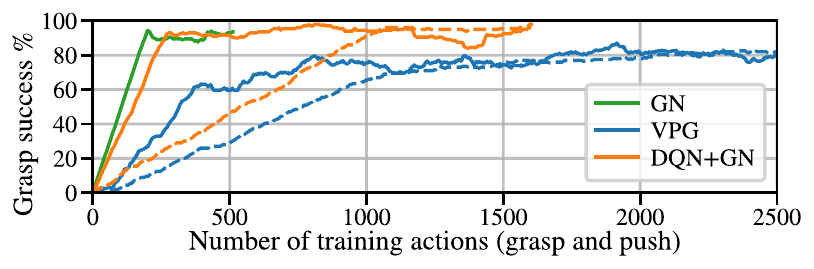}
    \caption{\label{fig:simulation-learning-curve}
    Grasp learning curves of algorithms for \pag in simulation. 
    The $x$-axis is the total number of training steps, i.e., number of actions taken, 
    including push and grasp. The $y$-axis is the grasp success rate.
    The dashed lines denote the success rate for a grasp right after a push action. 
    }
\end{figure}

The learning curve in simulation is provided in Fig.~\ref{fig:simulation-learning-curve}. 
The pre-training process (Alg.~\ref{alg:train}, line~\ref{alg:train-pre}, which is also 
self-supervised) for \gn takes $100$ offline images that are not reported in the plot. 
Comparing DQN+\gn which reaches $>90\%$ success rate with less than $300$ (grasp and push) 
samples, and baseline VPG, which converges at $~82\%$ success rate with more than $2000$ 
(grasp and push) samples, it is clear that \gn has significantly higher grasp success 
rate and sample efficiency than the baseline VPG.
As shown by the comparison between \gn and DQN+\gn, when training using only grasp 
actions, \gn can be more sample efficient without sacrificing success rate. 
The result also indicates that for randomly generated \pag, pushing is mostly 
\emph{unnecessary}.
 
\subsection{Evaluation of the Complete Pipeline}

We evaluate the learned policies on \pag with up to $30$ objects, first 
in a simulation, then on a real system. 
Four algorithms are tested: VPG~\cite{zeng2018learning}, DQN+\gn, REA+\dipn, and 
\dipngn (our full pipeline). 
Here, REA+\dipn follows Alg.~\ref{alg:evaluation} but uses the reactive grasp 
network from \cite{zeng2018learning} instead of \gn for grasp reward estimation.
We use three metrics for comparison: 
{\em (i) Completion}, calculated as {\em the number of \pag instances where all 
objects got removed} divided by {\em the total number of instances}; 
incomplete tasks are normally due to pushing objects out of the workspace, or 
cannot finish a task within a number of actions less than three times the number 
of objects. 
{\em (ii) Grasp success}, calculated as {\em the total number of objects grasped} 
divided by the {\em the total number of grasp actions}. 
{\em (iii) Action efficiency}, calculated as {\em the total number of objects 
removed} divided by {\em the total number of actions (grasp and push)}. 
For grasp success rate and action efficiency, we use two formulations: 
one does not count incomplete tasks (reported in gray text), which is the same as 
the one used in \cite{zeng2018learning}, and the other one counts incomplete tasks, 
which we believe is more reflective. The robot could grasp more than one object at a time. 
We consider it as a successful grasp, as the goal is to clear all objects from the table.

Table.~\ref{table:pag-sim} reports simulation results on $30$ randomly generated 
\pag instances and $10$ manually placed hard instances (illustrated in 
Fig.~\ref{fig:evaluation-hard-instances}) where push actions are necessary. 
The algorithms do not see the hard instances in training. 
The number of training samples for each algorithm are: 
$2500$ actions (grasp and push) for VPG~\cite{zeng2018learning}, 
$1500$ grasp actions and $2000$ push actions for REA+\dipn, 
$1500$ actions (grasp and push) for DQN+\gn,
and $500$ grasp actions and $1500$ push actions for \dipngn (our full pipeline). 
The results show that \dipn and \gn both are sample efficient in 
comparison with the baseline and provide significant improvement 
in \pag metrics; when combined, \dipngn reaches the highest performance 
on all metrics.

\begin{table}[ht!]
    \centering
    \caption{\label{table:pag-sim} Simulation, random and hard instances (mean $\%$)}
    \begin{tabular}{l|l|l|p{0.75cm}|p{0.75cm}|p{0.9cm}|p{0.9cm}}
    \hline
    & Method & Completion & \multicolumn{2}{c|}{Grasp success} & \multicolumn{2}{c}{Action efficiency} \\ \hline \hline
    \multirow{4}{*}{\;Rand\;\;}
    & VPG\cite{zeng2018learning}            & 20.0\footnotemark          
        & \textcolor{gray}{69.0}            & 52.6          
        & \textcolor{gray}{66.3}            & 52.6             \\ \cline{2-7}
    & REA\cite{zeng2018learning}+\dipn      & 83.3          
        & \textcolor{gray}{79.5}            & 77.9           
        & \textcolor{gray}{77.4}            & 76.3            \\ \cline{2-7}
    & DQN\cite{zeng2018learning}+\gn        & 46.7          
        & \textcolor{gray}{85.2}            & 83.9           
        & \textcolor{gray}{83.4}            & 81.7             \\ \cline{2-7}
    & \dipngn                               & \textbf{83.3} 
        & \textcolor{gray}{\textbf{86.7}}   & \textbf{85.2} 
        & \textcolor{gray}{\textbf{84.4}}   & \textbf{83.3}    \\ \hline \hline
    \multirow{4}{*}{\;Hard\;\;}    
    & VPG\cite{zeng2018learning}            & 77.7          
        & \textcolor{gray}{67.4}            & 60.0          
        & \textcolor{gray}{60.8}            & 57.6             \\ \cline{2-7}
    & REA\cite{zeng2018learning}+\dipn      & 90.3          
        & \textcolor{gray}{81.5}            & 76.6           
        & \textcolor{gray}{64.7}            & 62.6             \\ \cline{2-7}
    & DQN\cite{zeng2018learning}+\gn        & 86.0          
        & \textcolor{gray}{91.1}            & 87.1           
        & \textcolor{gray}{70.2}            & 67.9             \\ \cline{2-7}
    & \dipngn                               & \textbf{100.0} 
        & \textcolor{gray}{\textbf{93.3}}   & \textbf{93.3} 
        & \textcolor{gray}{\textbf{74.4}}   & \textbf{74.4}          \\ \hline
    \end{tabular}
\end{table}
\footnotetext{{The low completion rate is primarily due to pushing objects outside of the workspace, which happens less often in hard cases. }}

\begin{figure}[ht!]
    \centering
    \includegraphics[width = \linewidth]{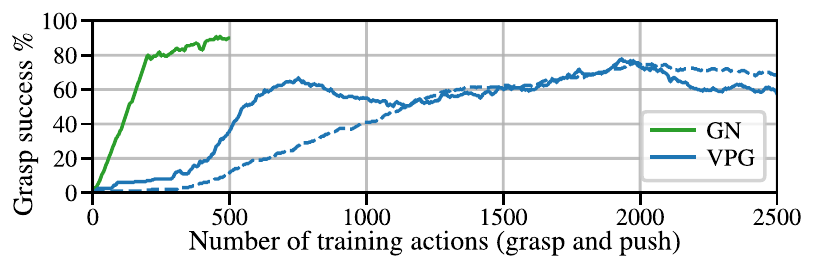}
    \caption{\label{fig:real-learning-curve} 
    Grasp learning curves for \pag in real experiment. 
    Solid lines indicate grasp success rate and dotted lines indicate push-then-grasp 
    success rates over training steps. The GN is trained in a grasp only manner.
    }
\end{figure}

We repeated the evaluation on a real system (see Fig.~\ref{fig:system-setup}).
Each random instance contains $10$ randomly selected objects; the hard instances are shown 
in Fig.~\ref{fig:evaluation-hard-instances}. 
Fig.~\ref{fig:real-learning-curve} shows grasp learning curve. 
We compare VPG~\cite{zeng2018learning} (trained with $2000$ grasp and push actions) and the 
proposed \dipngn pipeline (pre-trained with $100$ unlabeled RGB-D images for segmentation, 
trained \gn with $500$ grasp actions and \dipn with $1500$ simulated push actions). 
The evaluation result is reported in Table.~\ref{table:pag-real}. Remarkably, our networks, 
while being developed using only simulation based training, perform even better when 
trained/evaluated only on real hardware. 

\begin{table}[ht!]
    \centering
    \caption{\label{table:pag-real} 
    Real system, random and hard instances (mean $\%$)}
    \begin{tabular}{l|l|l|p{0.75cm}|p{0.75cm}|p{0.9cm}|p{0.9cm}}
    \hline
    & Method & Completion & \multicolumn{2}{c|}{Grasp success} & \multicolumn{2}{c}{Action efficiency} \\ \hline \hline
    \multirow{2}{*}{Rand\;}
    & VPG\cite{zeng2018learning}          & 80.0        
        & \textcolor{gray}{85.5}            & 79        
        & \textcolor{gray}{75.3}            & 67.9           \\ \cline{2-7}
    & \dipngn                             & \textbf{100.0} 
        & \textcolor{gray}{\textbf{94.0}}   & \textbf{94.0} 
        & \textcolor{gray}{\textbf{98.2}}   & \textbf{98.2}          \\ \hline \hline
    \multirow{2}{*}{Hard\;}
    & VPG\cite{zeng2018learning}          & 64.0        
        & \textcolor{gray}{75.1}            & 69.0         
        & \textcolor{gray}{51.9}            & 47.8           \\ \cline{2-7}
    & \dipngn  \;\;\;\;\;\;                           & \textbf{98.0\footnotemark} 
        & \textcolor{gray}{\textbf{89.9}}   & \textbf{89.9} 
        & \textcolor{gray}{\textbf{77.6}}   & \textbf{78.2}          \\ \hline
    \end{tabular}
\end{table}
\footnotetext{The single failure was due to an object that was successfully grasped 
but slipped out of the gripper before the transfer was complete.}

With \dipn and \gn outperforming the corresponding components from VPG~\cite{zeng2018learning}, 
it is unsurprising that \dipngn does much better. 
In particular, \dipn architecture allows it to learn intelligent, graded push behavior 
efficiently. In contrast, VPG~\cite{zeng2018learning} has a fixed $0.5$ push reward,
which sometimes negatively impacts the performance: VPG could push unnecessarily for 
many times without a grasp when it is not confident enough to grasp. It also risks 
pushing objects outside of the workspace.
Similar to \cite{zeng2018learning}, we also tested \dipngn with unseen objects, including 
soapboxes and plastic bottles. 
Our method obtained similar performance as reported in Table~\ref{table:pag-real}. 


\section{Conclusion}\label{sec:conclusion}
In this work, we have developed a Deep Interaction Prediction Network (\dipn) for learning 
to predict the complex interactions that occur as a robot manipulator pushes objects in 
clutter. Unlike most existing end-to-end techniques, \dipn is capable of generating accurate 
predictions in the form of clearly legible synthetic images that can be fed as inputs to a 
deep Grasp Network (\gn), which can then predict successes of future grasps. We demonstrated 
that \dipn, \gn, and \dipngn all have excellent sample efficiency and significantly 
outperform the previous state-of-the-art learning-based method for \pag challenges, while
using only a fraction of the interaction data used by the alternative. Our networks are 
trained in a fully self-supervised manner, without any manual labeling or human inputs, 
and exhibit high levels of generalizability. The proposed system was initially developed 
for a simulator, but it surprisingly performed better when trained directly on real 
hardware and objects. \dipngn is also highly robust with respect to changes in objects' 
physical properties, such as shape, size, color, and friction. In a future work, we will 
consider more complex manipulation actions, such as non-horizontal push actions and 
non-vertical grasps with arbitrary 6D end-effector poses, which are necessary for 
manipulating everyday objects in clutter. 


\bibliographystyle{IEEETran}
\bibliography{references}

\end{document}

%% file: related.tex
\noindent {\bf Grasping:} Robotic grasping techniques can generally be categorized in two main categories: {\it analytical} and {\it data-driven}~\cite{10.1109/TRO.2013.2289018}. 
Until the last decade, the majority of grasping techniques required precise analytical and 3D models of objects in order to predict the stability of {\it force-closure} or {\it form-closure} grasps~\cite{grasping,liang2019pointnetgpd,doi:10.1177/0278364912442972}. However, building accurate models
of new objects is a challenging task that requires thoroughly scanning all the objects in advance. Moreover, important material properties such as mass and friction coefficients are generally unknown. Statistical learning of grasp actions is an alternative data-driven approach that has received increased attention in the recent years. Most data-driven methods focused on isolated objects~\cite{DBLP:conf/iros/BoulariasKP11,Detry2013,Lenz2013,7139793,Yan-2018-113286,DBLP:conf/iccv/MousavianEF19}. Learning to grasp in cluttered scenes was explored in recent works~\cite{DBLP:conf/aaai/BoulariasBS14,Pas2015UsingGT,DBLP:conf/icra/PintoG16,pmlr-v78-mahler17a,mahler2017dexnet,kalashnikov2018qtopt}.
For example, a convolutional neural network was trained in the popular work~\cite{Pas2015UsingGT} to detect 6D grasping poses in point clouds. Such an approach suffers when multiple objects are mistaken for a single object. Another drawback is that using only grasp actions may be insufficient for handling dense clutter, making {\it pre-grasp} push actions necessary. 

\noindent {\bf Pushing:} Pushing is an example of non-prehensile actions that robots can apply on objects. As with grasping, there are two main categories of methods that predict the effect of a push action~\cite{10.3389/frobt.2020.00008}. \emph{Analytical} methods rely on mechanical and geometric models of the objects and utilize physics simulations to predict the motion of an object~\cite{Lynch-1993-15932,Mason86,Mason96,Mason99,Mason98,Mason-1985-15649}. 
Notably, Mason~\cite{Mason86} derived the voting theorem to predict the rotation and translation of an object pushed by a point contact.
A stable pushing technique when objects remain in contact was
also proposed in~\cite{Mason96}. 
These methods often make strong assumptions such as quasi-static motion and uniform friction coefficients and mass distributions. 
To deal with non-uniform frictions, a regression method was proposed in~\cite{Yoshikawa1991IndentificationOT} for identifying the support points of a pushed object by dividing the support surface into a grid.
The {\it limit surface} plays a crucial role in the mechanical models of pushing. It is a convex set of all friction forces and torques that can be applied on an object in quasi-static pushing. The limit surface is often approximated as an ellipsoid~\cite{doi:10.1177/027836499601500603}, or a higher-order convex polynomial~\cite{Zhou2016,JJZhou2018,DBLP:journals/ijrr/ZhouHM19}. An ellipsoid approximation was also used to simulate the motion of a pushed object to perform a push-grasp~\cite{DBLP:conf/rss/DogarS11}. To overcome the rigid assumptions of analytical methods, {\it statistical learning} techniques predict how new objects behave under various pushing forces by generalizing observed motions in training examples. For example, a Gaussian process was used to solve this problem in~\cite{DBLP:conf/icra/BauzaR17}, but was limited to isolated single objects. Most recent push prediction techniques rely on deep learning~\cite{NIPS2016_6161,7989023,watters2017visual}, which can capture a wider range of physical interactions from vision.
Deep RL was also used for learning pushing strategies from images~\cite{sergey2015learning,10.5555/2946645.2946684,DBLP:conf/icra/FinnL17,DBLP:journals/corr/GhadirzadehMKB17}. Our approach differs from the previous ones in two important aspects. First, our approach learns to predict the simultaneous motions of multiple objects that collide with each other. Second, the push predictions are used to plan pushing directions that improve the performance of a clutter removal system.

\noindent {\bf Push-grasping:} 
A pre-grasp sliding manipulation technique that was recently developed~\cite{RSS2020Changkyu,Kaiyu2019,DBLP:conf/iros/SongB20,L4DC2020Changkyu,DBLP:conf/aaai/BoulariasBS15} performs a sequence of non-prehensile actions such as side-pushing and top-sliding to facilitate grasping. 
Most existing methods for pre-grasp push require the existence of predefined geometric and mechanical models of the target object~\cite{Dogar2010PushgraspingWD,6631288,Dogar_2012_7076,King2013PregraspMA}. In contrast to these techniques, our approach does not require models of the objects. The present work improves upon the closely related Visual Pushing and Grasping (VPG) technique~\cite{zeng2018learning} notably by explicitly learning to predict how pushed objects move, and replacing model-free estimates of the Q-values of pushing actions with one-step lookaheads.